\documentclass[preprint,12pt]{elsarticle}

\usepackage{natbib}
\usepackage{hyperref}
\usepackage{enumitem}
\usepackage{booktabs, multirow} 
\usepackage{soul}
\usepackage[table]{xcolor} 
\usepackage{changepage,threeparttable} 
\usepackage{caption}
\usepackage{subcaption}
\usepackage{adjustbox}
\usepackage{algorithm}
\usepackage{epstopdf}

\usepackage{algpseudocode}
\usepackage{tikz}
\usepackage{pgfplots}
\usepackage{pgfplotstable}
\pgfplotsset{compat=1.7}
\usetikzlibrary{arrows.meta,
                chains,
                positioning}
\usepackage{amsmath}
\algrenewcommand\algorithmicensure{\textbf{}}
\def\tsc#1{\csdef{#1}{\textsc{\lowercase{#1}}\xspace}}
\tsc{WGM}
\tsc{QE}
\tsc{EP}
\tsc{PMS}
\tsc{BEC}
\tsc{DE}

\begin{document}
\let\WriteBookmarks\relax
\def\floatpagepagefraction{1}
\def\textpagefraction{.001}



\title{FedStack: Personalized Activity Monitoring using Stacked Federated Learning}                   



%
\author[1]{Thanveer Shaik\corref{cor1}}
\ead{thanveer.shaik@usq.edu.au}


\address[1]{School of Mathematics, Physics and Computing, University of Southern Queensland, Toowoomba 4350, Australia}

\author[1]{Xiaohui Tao\corref{cor2}}
\ead{Xiaohui.Tao@usq.edu.au}


\author[2,3]{Niall Higgins}

\ead{Niall.Higgins@health.qld.gov.au}


\address[2]{Metro North Hospital and Health Service, Royal Brisbane and Women's Hospital, Herston 4029, Australia}
\address[3]{School of Nursing, Queensland University of Technology, Brisbane 4000, Australia}

\author[4]{Raj Gururajan}
\ead{Raj.Gururajan@usq.edu.au}

\author[5]{Yuefeng Li}
\ead{y2.li@qut.edu.au}

\address[5]{School of Computer Science, Queensland University of Technology, Brisbane, Australia}

\author[4]{Xujuan Zhou}
\ead{Xujuan.Zhou@usq.edu.au}


\address[4]{School of Business, University of Southern Queensland, Springfield 4300, Australia}

\author[7]{U Rajendra Acharya}
\ead{Rajendra_Udyavara_ACHARYA@np.edu.sg}
\address[7]{Singapore University of Social Sciences, Singapore}



\begin{abstract}
Recent advances in remote patient monitoring (RPM) systems can recognize various human activities to measure vital signs, including subtle motions from superficial vessels. There is a growing interest in applying artificial intelligence (AI) to this area of healthcare by addressing known limitations and challenges such as predicting and classifying vital signs and physical movements, which are considered crucial tasks. Federated learning is a relatively new AI technique designed to enhance data privacy by decentralizing traditional machine learning modeling. However, traditional federated learning requires identical architectural models to be trained across the local clients and global servers. This limits global model architecture due to the lack of local models' heterogeneity. To overcome this, a novel federated learning architecture, FedStack, which supports ensembling heterogeneous architectural client models was proposed in this study. This work offers a protected privacy system for hospitalized in-patients in a decentralized approach and identifies optimum sensor placement. The proposed architecture was applied to a mobile health sensor benchmark dataset from 10 different subjects to classify 12 routine activities. Three AI models, artificial neural network (ANN), convolutional neural network (CNN), and bidirectional long short-term memory (Bi-LSTM) were trained on individual subject data. The federated learning architecture was applied to these models to build local and global models capable of state-of-the-art performances. The local CNN model outperformed ANN and Bi-LSTM models on each subject data. Our proposed work has demonstrated better performance for heterogeneous stacking of the local models compared to homogeneous stacking. Further analysis of the global heterogeneous CNN model determined that the optimum placement of the sensors on human limbs resulted in better activity recognition. This work sets the stage to build an enhanced RPM system that incorporates client privacy to assist with clinical observations for patients in an acute mental health facility and ultimately help to prevent unexpected death.
\end{abstract}


\begin{highlights}
\item A novel federated architecture, FedStack, is proposed to overcome the heterogeneity limitation in traditional federated learning.
\item Enhanced personalized patient monitoring by adopting the proposed novel federated architecture to classify physical activities.
\item FedStack framework outperformed the baseline models' performance in federated learning.
\end{highlights}


\maketitle

\section{Introduction}
%
%
%
%
Remote patient monitoring (RPM) is a trending application in health intelligence to identify health parameters using sensors without obstructing a person's day-to-day activities. Typical RPM systems can track and record vital signs such as heart rate and breathing rate, but they can also be applied to measuring physical activities like walking, running, unintentional falls, and so on. This is achieved through a wide variety of device applications like smartwatches~\cite{seshadri2020wearable}, smart shirts~\cite{wu2019wearable}, telehealth~\cite{Lafta2016} and mobile sensors~\cite{ramkumar2019remote}\cite{uddin2020body}.

Artificial intelligence (AI) is being used in a variety of health applications~\cite{Chen2022} as such as image processing~\cite{ClassPeters2021,Blais2022}, natural language processing~\cite{9781308}, sensor data processing~\cite{Sofi2022}, and so on. The use of artificial intelligence (AI) can enhance the capabilities of RPM systems through processing the recorded data and by training deep machine learning models to build efficient predictive systems. An example of this is the use of early warning scores (EWS) that have been designed by clinicians to detect early signs of patient deterioration. Typical RPM systems predict possible future clinical events based on recorded data as well as real-time time-series data. These assistive applications can be particularly useful for acute inpatient care, but they can also be applied to those being cared for in their home as a strategy to manage the current pandemic. An important consideration when designing RPM systems is to ensure the confidentiality of health information and be able to adapt to the business processes of clinical activities. Current research approaches promote homogeneous data-centric models built on a centralized data server. This method of generalizing the data learning could limit the application of RPMs to health care that needs to be person-centric and individualized. 

\begin{figure}[ht]
    \centering
    \includegraphics[width=\columnwidth]{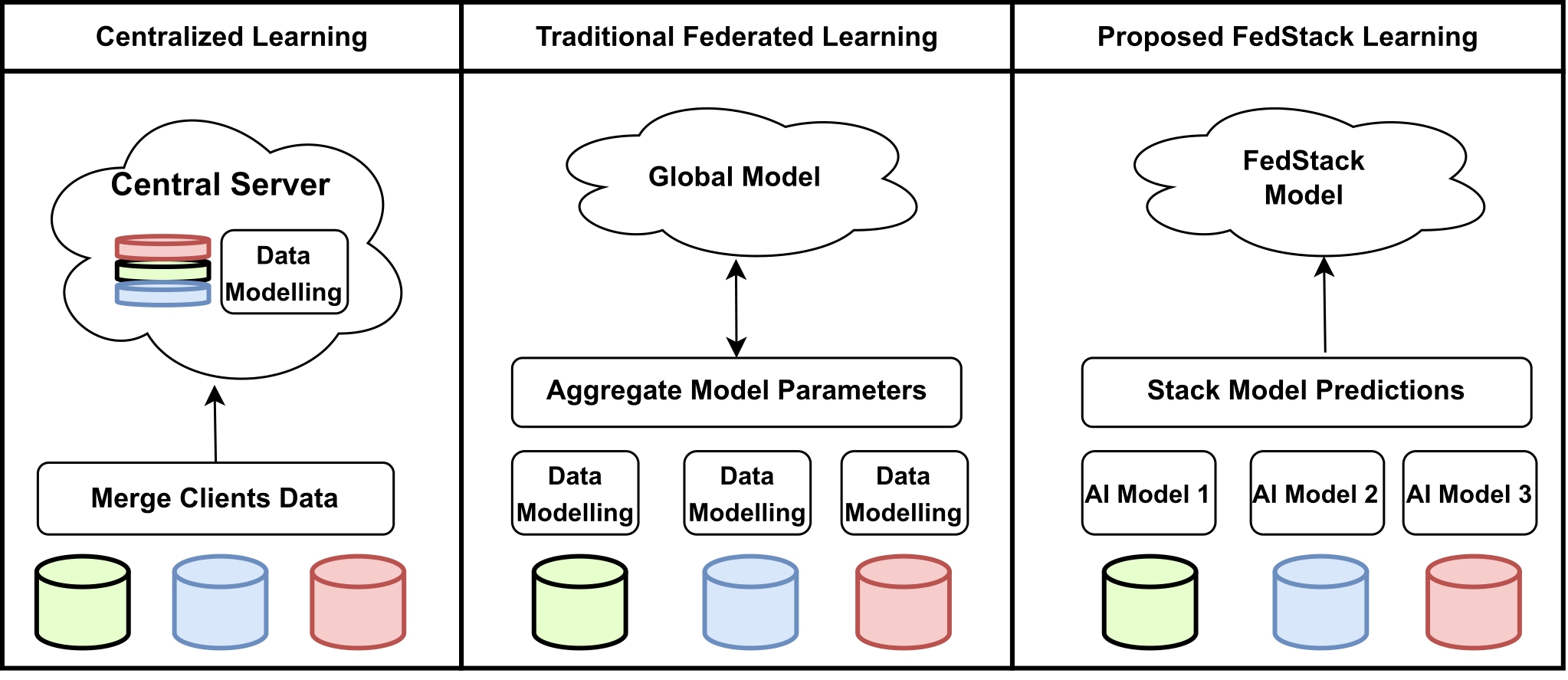}
    \caption{Research Background}
    \label{fig:intro}
\end{figure}

In healthcare applications, each patient has different demographics and health history, which requires personalized monitoring. Traditional centralized learning approaches merge all client data to a cloud server and host a model as shown in Fig.~\ref{fig:intro}. A centralized architecture cannot cater to the needs of personalized monitoring and compromises client privacy. Decentralized learning can focus on individual client data, and this can be achieved using a recently developed method called federated learning (FL)~\cite{bonawitz2021federated} as shown in Fig.~\ref{fig:intro}. This could overcome the privacy issues of centralized learning. However, it has a limitation of aggregating heterogeneous architectural client models. 

The research problem is local clients are compelled to use similar architectural models as part of their data modeling in traditional FL, which might be impractical as each client might have different requirements and priorities. A heterogeneous stacked federated learning architecture, FedStack is proposed to address this problem. The primary aim of the proposed framework is to decentralize the machine learning approach by allowing each device or client to train a machine learning algorithm on their private data locally. Upon evaluation, the trained model predictions are communicated to a global model residing in a separate server, thus decentralizing the client monitoring process and preserving their privacy. The global model would then retrieve stacked predictions from different devices or clients and update the central machine learning model using heterogeneous data. The secondary aim of this study is to adopt the framework of an RPM system to isolate each set of patient data, protect their privacy, and train AI models locally. The proposed FedStack also determines the optimum positions to place a sensor on a human body to achieve greater activity recognition capability.

This study used a benchmark dataset with tri-axial sensor data collected from 10 healthy volunteer subjects. Sensor data from each subject was fed to three different AI models to classify and evaluate their activities. The proposed FedStack learning approach was used to isolate each subject's data and ensemble the predictions of individual subject models. The ensemble predictions were then communicated to a global model. Three different AI models Artificial Neural Network (ANN), Convolutional Neural Network (CNN), and bidirectional long short-term memory (Bi-LSTM) were trained as global and local models on each subject data, and their classification performances were evaluated and compared. The three AI models ANN, CNN, and Bi-LSTM on local client data achieved an average balanced accuracy of 0.98, 0.99, and 0.93 respectively. CNN model has outperformed the other two AI models on all nine subjects’ data. The predictions from the local client models were ensembled and trained in the global model. The global CNN model has outperformed the other two models with a balanced accuracy of 0.976 and 0.996 for homogeneous and heterogeneous stacked predictions respectively. The global heterogeneous CNN model was evaluated with one sensor input at a time to determine the optimum positions to place a sensor on a human body. The sensors on the right wrist and the left ankle were optimum sensor positions for human activity recognition. The global CNN model with the right wrist and left ankle sensors data achieved balanced accuracy of 0.99 and 0.99 respectively.

The ultimate goal of this research is to detect accurate vital signs and natural body movements of multiple mobile patients in an acute mental health setting. As part of this research, a simulated psychiatric hospital ward was established using a remote patient monitoring (RPM) system utilizing sensors and radio frequency identification (RFID) technology. Optimum positions of RFID reader-antennas were determined in the simulated ward based on received signal strength indicators (RSSI) from passive RFID tags ~\cite{tao2021remote}. Signals detected were considered vital signs originating from subtle motions from the patient's body, and those from larger body movements were considered indicative of physical activities. This research offers a method to classify physical activities using AI models and compares their performances. FL is introduced to protect individual patient privacy and enhance the AI architecture with decentralized modeling. The proposed approach was able to classify the labels and outperform the state-of-art works in each local model and global model. The study contributions are as follows:

\begin{itemize}
    \item This study proposed a novel heterogeneous stacked federated learning architecture to overcome the limitation of heterogeneous architectural ensembling in the traditional federated learning approach.
    \item This research combines tri-axial data of multiple sensors on the human body to track their natural body movements using federated learning in the area of healthcare.
    \item The proposed approach achieved better accuracy than current baseline models for human activity recognition by using AI models.
    \item In this study, Federated learning is introduced at a subject level to train an AI model with individual subject data and design a personalized monitoring system.
    \item This study determines the optimum placement of sensors on the human body for activity recognition based on individual sensor data contribution in classifying the label activities.
\end{itemize}

Section~\ref{litearture} presents related works on human activity recognition (HAR) using traditional machine learning methods, DL methods, and FL methods. Section~\ref{methodology} presents the research problem formulation, FedStack architecture proposed in this study, and the methodology of adopting the proposed framework for personalized patient monitoring. Experimental design, baseline models, and performance metrics are discussed in Section~\ref{exp-design}. In Section~\ref{results}, experiment results and their analysis, baseline models comparison, and discussion on proposed research results have been presented. Finally, the paper concludes in Section~\ref{conclusion}.


\section{Related Works}\label{litearture}

\subsection{Traditional Machine Learning Methods}
Sri Harsha et al.~\cite{sri2021performance} analyzed commonly used machine learning algorithms for sensor-based human activity recognition. The authors built a HAR system based on tri-axial accelerometer and gyroscope data collected via mobile phones. The data were classified into running, walking, climbing up or down activities using support vector machines (SVM), decision trees, and random forest models. These algorithms were evaluated using the Gini index, and random forest outperformed the other models in classifying the running or walking with an accuracy of 94.43\%. Overall, the models achieved moderate accuracy of 63.68\%, 63.83\%, and 68.07\% for SVM, decision trees, and random forest, respectively. Halim~\cite{Halim2022} proposed stochastic recognition of personalized human daily activity recognition using hybrid descriptors and random forests. Erhan et al.~\cite{bulbul2018human} classified activities of walking, climbing up or down, sitting, standing, and laying down with accelerometer and gyroscope data collected from a smartphone. The authors used machine learning models, namely decision trees, SVM, K-nearest neighbors (KNN), and ensemble classification methods boosting, bagging, and stacking. The SVM model achieved the highest accuracy of 99.4\% compared to the other classification models. Asim et al.~\cite{asim2020context} presented interesting work with a novel framework designed for HAR. The authors incorporated human behavioral contexts in activity recognition. Six different context-independent activities of lying down, standing, bicycling, sitting, running, and standing along with 15 different behavioral contexts were chosen as primary activities for recognition originally described by Vaizman et al.~\cite{vaizman2017recognizing}. These activities were classified using decision trees, KNN, SVM, random forest, and Naive Bayes classifiers. The authors compared the classifier's performance for context-independent HAR to context-dependent HAR. The random forest classifier performed better in classifying both sets of activities.

\subsection{Deep Learning Methods}
DL methods have expanded their scope in diverse applications like predicting traffic flow. Wang et al.~\cite{Wang2021} proposed Multitask Recurrent Graph Convolutional Network (MRGCN) to predict traffic flows accurately in a city. Essien et al.~\cite{Essien2020} bidirectional long short-term memory stacked autoencoder to predict traffic flow from tweet messages with traffic and weather information. DL methods overcome challenges traditional AI models face by learning efficient features from raw sensor data and customizing a hierarchy from low-level features to high-level abstractions. Moreover, DL methods can extract features automatically in a task-dependent manner~\cite{murad2017deep}. Murad and Pyun~\cite{murad2017deep} use five public HAR datasets to compare the performance of deep recurrent neural networks (DRNNs) with conventional mechanical methods like SVM, random forest, and KNN. The authors presented unidirectional, bidirectional, and cascaded architectures on long short-term memory (LSTM) DRNNs and found that unidirectional DRNN on the USC-HAD dataset~\cite{zhang2012usc} had the highest accuracy of 97.8\%. Suto et al.~\cite{suto2018comparison} also tested the efficiency of DL on real-time data collected through self-learning activity recognition applications. The authors classified activities such as cycling, running, jogging, walking, sitting, standing, and lying using a convolutional neural network (CNN), artificial neural network (ANN), and 1NN. CNN achieved an accuracy of 94.2\%, but its long training time was a limiting factor for their usage in real-time HAR applications. Instead, the authors opted for a well-constructed ANN to obtain optimal results.

\subsection{Federated Learning}
An increase in electronic assistive health applications like smartwatches and activity trackers led to pervasive computing or ubiquitous computing where each device can seamlessly exchange data with another~\cite{alam2020fog}. Although it has the advantage of tracking real-time changes in personalized human health data being centralized for monitoring, it is vulnerable to security breaches of data privacy~\cite{dang2019survey}. As AI has matured, a vast amount of human data is being generated worldwide. To manage this huge data, technology company Google introduced a mechanism that trains a machine learning algorithm across multiple decentralized devices or servers without exchanging their local data samples and focusing on personalized data management. This is called federated learning (FL) also known as collaborative learning~\cite{bonawitz2021federated}. FL overcomes the issues of data privacy that exist with traditional centralized learning techniques where all device or server data is merged for analysis. Federated Learning has been widely adopted by various applications such as semi-supervised credit prediction~\cite{Li2022}.

Currently, personalized human activity recognition is achieved using cloud-based traditional machine learning algorithms and DL algorithms. FL enables on-device training and shares its model parameters to be aggregated instead of the server's global model. Sannara et al.~\cite{ek2020evaluation} evaluated the performance of FL aggregation techniques like FedAvg, FedMa, and FedPer against centralized training techniques. The CNN model was used to classify eight physical activities. However, even though the FL techniques outperformed the local client models, the server model accuracy of the FL techniques was low compared to centralized learning. An activity recognition system was designed by Zhao et al.~\cite{zhao2020semi} that was based on semi-supervised FL. The authors used unsupervised learning on clients to update LSTM autoencoders locally and then communicate the models to a global server that executed supervised learning using softmax classifiers. The activity recognition system was built using LSTM to process the time series data. A personalized human activity recognition system based on the FederatedAveraging method, HARFLS, was proposed by Xiao et al.\cite{xiao2021federated}. The authors stressed the need for feature extraction and designed a perceptive extraction network based on a convolutional block. HARFLS, with the extractive network, was able to outperform existing human activity recognition works. Another personalized indoor activity recognition system was based on Federated Markov Logic Network (FMLN) framework developed by Zhang et al.~\cite{zhang2022federated}. Xiaomin et al.~\cite{ouyang2021clusterfl} proposed a slightly different clusterFL approach to minimize the empirical loss of trained models by exploiting the similarity of users' data and improving federated model accuracy and communication efficiency between local models and global models.

\subsection{Summary}

Traditional machine learning, DL, and FL are subsets of AI with different approaches toward human activity recognition. Traditional machine learning models can classify human activities but need domain expertise to reduce data complexity. This problem can be addressed using DL models to apply an iterative approach to processing data and learning features. However, neither traditional machine learning nor DL methodologies can safeguard data privacy without a supportive framework. The need for DL to rely on huge data for better modeling results led to centralized data management systems, hindering personalized monitoring. FL is a framework designed to address these issues by decentralizing the modeling architecture. In this study, a novel heterogeneous FL architecture was designed with three AI models consisting of a machine learning model and two DL models both locally and globally to not only take advantage of the robust learning of DL models but also to build a personalized monitoring system using decentralized federated architecture. Client privacy can be protected due to federated learning characteristics. The proposed design overcomes the limitation of heterogeneous architectural ensembling of local client models in traditional federated learning.

\begin{figure*}[ht]
    \centering
    \includegraphics[width=\columnwidth]{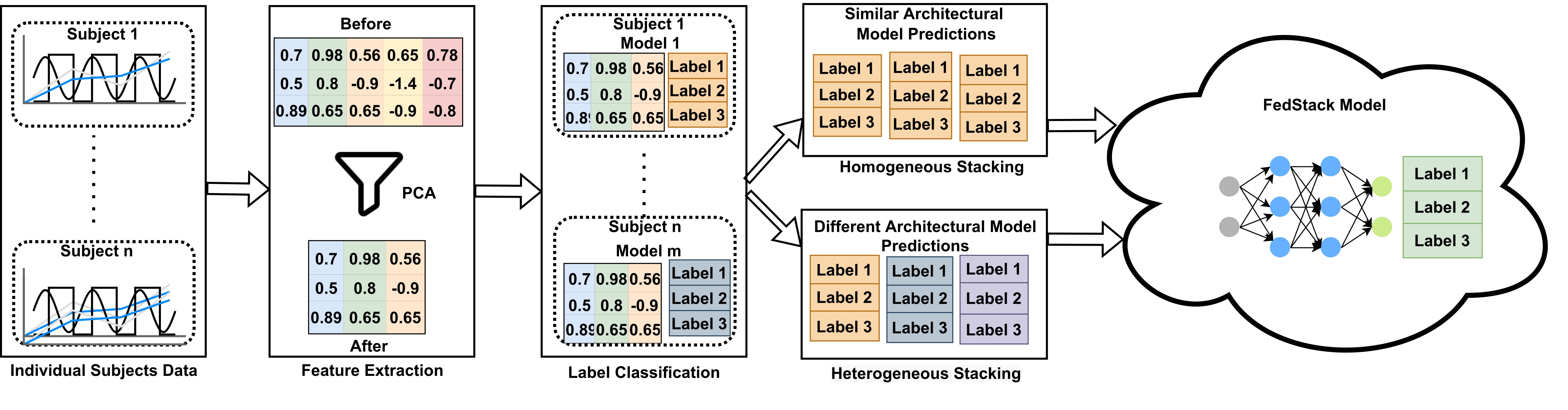}
    \caption{Proposed FedStack architecture.}
    \label{fig:federated_architecture}
\end{figure*}

\section{Methodology}\label{methodology}

\subsection{Research Problem Formulation}\label{problem}
The primary aim of this study is to design a heterogeneous FL process to build a global model across a variety of AI architectural models that are trained at individual clients or devices. Irrespective of data distribution or model architecture at an individual client, a robust heterogeneous global FL model needs to be designed. For Example, let's say $n$ clients or their devices are using $m$ different AI models for local data modeling of their data, and each client estimated predictions $p$. The objective is to pass the predictions $m$ while holding the local data at the client level or local device. The problem can be mathematically defined in Equation~\ref{fedearted_equation}.

\begin{equation}\label{fedearted_equation}
    {Train(G) \longleftarrow {\sum _{i=1}^{n} m_{i}(p_{i})}}
\end{equation}
where:
\begin{itemize}
    \item $Train(G)$: Train Global Model ($G$) with local model predictions
    \item $m_{i}(p_{i})$: Local model($L$) with their predictions ($p$) at each client $i=1,2,3..n$
\end{itemize}

The secondary aim of this research is to adopt the heterogeneous FL process to personalize patients' physical activity monitoring while protecting their private data and classify the activities on three-dimensional sensor data. The approach will also determine the optimum position of sensors on the human body that will classify human activity appropriately. This is achieved by analyzing each sensor data recorded from different human body positions in classifying their physical activities.

\subsection{FedStack Learning Framework}
In this study, a novel federated learning framework, FedStack is proposed to build a heterogeneous global FL model by stacking predictions of individual client models. This approach allows heterogeneous architectural models at the client level, overcoming the issue of heterogeneity~\cite{li2020federated} in traditional FL techniques. Let's assume that $n$ number of devices with different AI models are getting trained for analytical purposes. For example, One client might use neural networks, others might use deep learning models or even linear models like generalized linear models (GLM), and so on. Each of these AI models has different architectural structures~\cite{valizadeh-parde-2022-ai} and traditional FL techniques such as FedAvg cannot aggregate due to different internal and external parameters for each architectural structure.

Let's say three different clients with three distinct architectural AI models are being trained with their local private data to predict or classify labels. The predictions of the three client models can be estimated in Equations~\ref{eq:linear},\ref{eq:ann_node},\ref{eq:ann_out},\ref{eq:cnn_out}. Equation~\ref{eq:linear} is from a linear regression model used by the first client, in which response and input features are assumed to have a linear relationship. The $y_{i}$ is predicted response variable for individual observations $i$, $\beta_{j}$ are coefficient of each input features $x_{i}$, $\epsilon_{i}$ are random errors. The second client uses a non-linear model with three layers, an input layer, a hidden layer, and an output layer. Equation~\ref{eq:ann_node} presents the output of each node or neuron in a layer, which is configured with weight $w_{i}$ and bias $b$ for each input $x$ and $z$ output of the node. The output will be passed to a next-layer neuron for processing. Equation~\ref{eq:ann_out} presents three layered output combined to outcome the prediction $y$. The third client uses convolution neural networks to model and analyze their private information and is represented by Equation~\ref{eq:cnn_out}.

\begin{equation}\label{eq:linear}
    \displaystyle{y_{1i} = \beta_{0}+\beta_{1}x_{i}+...+\beta_{p}x_{p}+\epsilon_{i}}
\end{equation}

\begin{equation}\label{eq:ann_node}
    \displaystyle{z=f(b+x.w)=f(b+\sum _{i=1}^{n} x_{i}w_{i})}
\end{equation}
\begin{equation}\label{eq:ann_out}
     \displaystyle{y_{2i}=f(f(f(x.w_{1}).w_{2}).w_{3})}
\end{equation}

\begin{equation}\label{eq:cnn_out}
    \displaystyle{y_{3i}=b_{i} + \sum _{c=0}^{n_{c}-1} \sum _{k=-p}{p} x_{c,j-k}w_{c,k}}
\end{equation}

These three AI models have different architectures and configurations for three different clients or devices. Traditional FL techniques aggregate the local model architectures to build a robust global model, but they have a limitation in aggregating the heterogeneous model. The proposed FedStack framework can overcome limitations and build a heterogeneous global FedStack model across heterogeneous devices with different AI models. This can be achieved by heterogeneous stacking (non-identical architectural models), in which the predictions of different architectural models are stacked as shown in Equation~\ref{eq:het_stack}. The heterogeneous stacked predictions of the three models $y_{1i},y_{2i},y_{3i}$, where ${1i},{2i},{3i}$ denotes non-identical architectural models. These predictions are derived from Equations~\ref{eq:linear},\ref{eq:ann_out},\ref{eq:cnn_out} for heterogeneous stacking and used to train the global FedStack model as shown in Fig.~\ref{fig:federated_architecture}. The framework is designed to support the aggregation of identical architectural models, but with the stacking predictions, not an average of model weights (FedAvg). This process is called homogeneous stacking, as shown in Equation~\ref{eq:hom_stack}. The predictions $y_{1i}$ of similar architectural models from different are stacked to train the global model. To show identical architectural model predictions, ${1i}$ was used in Equation~\ref{eq:hom_stack}.

\begin{equation}\label{eq:het_stack}
    \displaystyle{Train(G) \longleftarrow stack(y_{1i},y_{2i},y_{3i})}
\end{equation}

\begin{equation}\label{eq:hom_stack}
    \displaystyle{Train(G) \longleftarrow stack(y_{1i},y_{1i},y_{1i})}
\end{equation}

\begin{figure*}[!ht]
    \centering
    \includegraphics[width=\columnwidth]{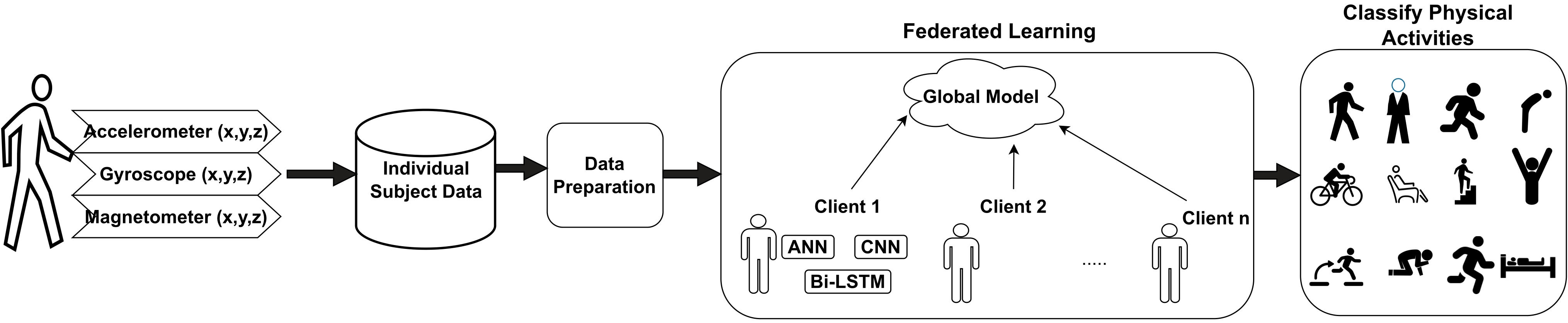}
    \caption{Research architecture overview.}
    \label{fig:framework}
\end{figure*}

\subsection{Personalized Patient Monitoring}

FedStack proposed in this study adopted the novel federated learning framework to analyze the individual subject or device data and build a global model by aggregating individually trained local AI models. All parameters of the local model's predictions were aggregated using the traditional stacking ensemble technique and compiled in the global model as shown in Equation~\ref{eq:het_stack},\ref{eq:hom_stack}. The predictions of each AI model on each subject data were stacked homogeneously and heterogeneously and passed to the global model. This enabled a heterogeneous architecture for federated learning, where clients could have a variety of model architectures. The proposed FL framework can be adapted by RPMs to monitor patients' activities based on classification models. The models' outcomes on individual patient data can stack to build the FedStack global model, as shown in Fig.~\ref{fig:framework}.


Traditional machine learning and DL models consolidate or centralize data on a single machine or a data server, requiring users' private data to interpret the results. The Federated Learning-based architecture shown in Fig.~\ref{fig:federated_architecture} has been designed for this research. This was introduced to decentralize the data training approach and avoid centralizing the data on one machine or a data center. It comprises local models and a global model. All local models were trained individually on user devices, including public and private data. Both traditional machine learning models and DL models are suitable for this architecture. Based on the training on individual data, the model parameters such as model predictions were communicated to the global model executed in a cloud server. Implementing global models in the cloud can run random rotations on local models and retrieve local model parameters or predictions without the need for the original user's data. Random rotations were executed based on the individual user or device data dimensions. Once the random devices or users were selected for the FL process, model predictions were stacked using the traditional ensemble method of stacking~\cite{9445049}. 

Traditional machine learning methods would require heuristic hand-crafted feature extraction to enhance their performance~\cite{alawneh2022personalized}. Specifically, personalized activity recognition would involve diverse data knowledge, including sensors, limiting traditional machine learning methods. In contrast, deep learning methods can learn features and automate model building~\cite{janiesch2021machine}. Although neural network family algorithms have been criticized for their black box nature~\cite{galan2019assessing}, deep learning models have been known for robust and efficient performance. The research community widely adopts these models for classification tasks related to human activity recognition~\cite{murad2017deep, zhang2012usc,suto2018comparison, zhao2020semi, ek2021federated, ouyang2021clusterfl, cho2018divide, ronao2016human, ronao2015deep, jiang2015human,almaslukh2017effective, ignatov2018real}. Three AI models, Artificial Neural Network (ANN), Convolutional Neural Network (CNN), and bidirectional long short-term memory (Bi-LSTM) were selected for this study because it does not need domain expertise and data complexity is greatly reduced. Each of these models has different architectures, which support the novelty claim of heterogeneous model training across different clients and training global models with heterogeneous model predictions. Three iterations were designed for this study, with one AI model trained as a local model and a global model in each iteration.
\subsubsection{Feature Reduction}
Feature reduction techniques were subsequently implemented on the independent variables to reduce the dimensionality and filter noise from the independent variables. A principal component analysis (PCA) was implemented as part of the dimensionality reduction for each subject dataset. The PCA created new uncorrelated variables called principal components, which maximized variance among the features by transforming the input data into a new coordinate system~\cite{tkachenko2018model}. The transformed variables were used to compute the covariance matrix, which led to calculating eigenvalues and corresponding eigenvectors for the matrix. Based on the cumulative sum of explained variance ratio retrieved from PCA, eigenvalues and the cumulative sum of the eigenvalues were computed. The principal intent of using PCA was to put the maximum possible information in the first components so that the latter could be excluded from the model training. With this, filtered noise was reduced, and feature reduction will be conducted. The reduced features were split into learning and testing data, and the learning set was used to train the AI models and evaluate their performance with the test set.

\subsubsection{Artificial Neural Network (ANN)}
ANN~\cite{russell2016artificial} is a collection of connected nodes called artificial neurons. Each neuron receives an input signal to a process and passes it on to the next layer of the ANN. A simple ANN can have only one input layer and one output layer called a single-layer network. It can extend to multiple layers, where hidden layers will be added between the input and output layers. The input layer of the ANN used in this study had nodes equal to the number of features selected, the output layer had nodes equal to the number of labels, and the hidden layers were invisible layers whose count depended on the prediction or classification complexity of the problem. Weights and biases were added to each hidden layer, and the transformed inputs were transmitted to the next layers with an activation function. Each neuron had input with weight and bias, as shown in Equation~\ref{neuron}. This simple ANN is mathematically represented in Equation~\ref{ann-eq} illustrating single input and output layers with activation functions to calculate weights, and bias on the input value. 

\begin{equation}\label{neuron} 
   y=f(b + \sum _{i=1}^{n} x_{i}.w_{i})
\end{equation}

\begin{equation}\label{ann-eq}
\begin{split}
y(x) = \sum_{i=1}^{n} Activation1(b+w_{i} x_{i}) \\
ANN(y) = Activation1(\frac{e^{y_{i}}}{\sum _{j=1}^{k} e^{y_{j}}})
\end{split}
\end{equation}

where:
\begin{itemize}
    \item $b$: Bias added on each hidden layer
    \item $x$: Input value.
    \item $w$: Weights added on each hidden layer
    \item $y$: Output value from each neuron 
    \item $Activation1$: Activation functions on input and hidden layers.
    \item $Activation2$: Activation function on output layer.
\end{itemize}

ANN can be executed with three layers including an input layer, a hidden layer, and an output layer~\cite{goh2021multimodal} with loss function binary cross-entropy from Keras. Rectified linear unit (ReLU) function has an activation function, and it has a limitation of defining negative inputs to zero, which deactivates the nodes or neurons. To overcome this challenge in datasets with negative attributes, leaky rectified linear unit (LeakyReLU)~\cite{9413590} activation was adopted. This is an extension of conventional ReLU activation which defines the negative inputs as an extremely small linear component as shown in Equation~\ref{LeakyReLU}. The function returns input value x as it is for all positive inputs, and for negative inputs returns a small value of $0.01*x$.

\begin{equation}\label{LeakyReLU}
    {f(x)=max(0.01*x,x)}
\end{equation}

Softmax activation was used to normalize the output into a probability distribution of classifying the record into one of the label activities. The threshold on the probability was then determined to transform values to label classification. 

\subsubsection{Convolutional Neural Network (CNN)}
CNN~\cite{9181233} is a DL model which was developed for image classification tasks where the 2-dimensional~(2-D) data can be interpreted. The CNN model is modified for human activity recognition by using 1-dimensional~(1-D) convolutional neural networks in each layer. Each input sensor signal is then read to prepare an internal representation of the input, so it can be mapped to an activity. Equation~\ref{cnn-eq} presents the mathematical notation of the 1-D CNN model with different activation functions adopted in each model design layer.

\begin{equation}\label{cnn-eq}
\begin{split}
y(x) = \sum_{i=1}^{n} Activation1(b+w_{i} x_{i})\\
y(x) = \sum_{i=1}^{n} Activation2(b+w_{i} x_{i})\\
CNN(y) = Activation3(\frac{e^{y_{i}}}{\sum _{j=1}^{k} e^{y_{j}}})
\end{split}
\end{equation}

where:
\begin{itemize}
    \item $b$: Bias added on each hidden layer
    \item $x$: Input value.
    \item $w$: Weights added on each hidden layer
    \item $y$: Output value from each neuron 
    \item $Activation1$: Activation functions on input and hidden layers.
    \item $Activation2$: Activation function on input and hidden layers.
    \item $Activation3$: Activation function on output layer.
\end{itemize}

MaxPool1D~\cite{ronald2021isplinception}, a pooling operation that the maximum value for a feature set and used to create a down-sampled group feature. Following a convolutional layer, the pooling operation was conducted as one of the layers. The pooled features were flattened~\cite{hamad2021dilated} into a 1-D array before processing in the output layer of the CNN model. The output layer provided a probability of each label classification, which was optimized using a threshold value to classify the features into a label.

\subsubsection{Bidirectional Long Short-Term Memory (LSTM)}
\begin{figure}[!ht]
    \centering
    \includegraphics[width=120mm,height=50mm]{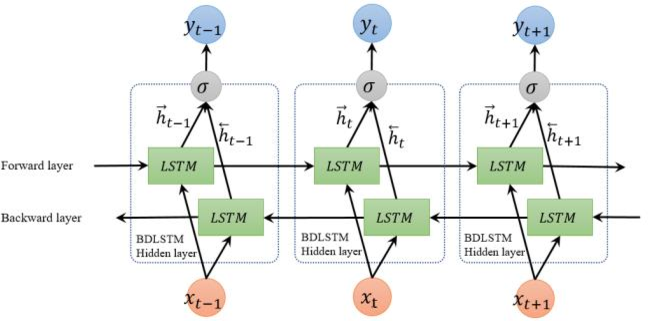}
    \caption{Bi-LSTM architecture~\cite{cui2020stacked}.}
    \label{fig:Bi-LSTM}
\end{figure}

LSTM model is a type of recurrent neural network (RNN) with a similar architecture. There are different variants of LSTM models like traditional, uni-directional, and bidirectional LSTM. Memory blocks act as the main component in the LSTM layer. There are three gates input, output, and forget gates for an LSTM block which denotes write, read and reset operations. The bidirectional LSTM cell state carries the information from past and future contexts to predict an element. Graphically, bidirectional LSTM is presented in Fig.~\ref{fig:Bi-LSTM}~\cite{cui2020stacked}. Mathematically, the Bi-LSTM model is defined in Equation~\ref{lstm-eq}. A regularization method, dropout~\cite{hassan2021end} was used to exclude activation and weight updates of recurrent connections from LSTM units probabilistically.

\begin{equation}\label{lstm-eq}
\begin{split}
y(x) = \sum_{i=1}^{n} Activation1(b+w_{i} x_{i})\\
Bi-LSTM(y) = Activation2(\frac{e^{y_{i}}}{\sum _{j=1}^{k} e^{y_{j}}})
\end{split}
\end{equation}

where:
\begin{itemize}
    \item $b$: Bias added on each hidden layer
    \item $x$: Input value.
    \item $w$: Weights added on each hidden layer
    \item $y$: Output value from each neuron 
    \item $Activation1$: Activation functions on input and hidden layers.
    \item $Activation2$: Activation function on output layer.
\end{itemize}
\begin{algorithm}
\small
\caption{Proposed stacked Federated learning algorithm}\label{alg:cap}
\begin{algorithmic}[1]
\Ensure{}{\textbf{Input:} {a set of subjects $\mathcal{C}=\{1,2,\dots,C\}$}; {a set of AI models $\mathcal{M}=\{1,2,\dots,M\}$}; {a set of labels $\mathcal{K}=\{1,2,\dots,K\}$}\vfill}
\Ensure{}\textbf{Output:} Classification probabilities of $\mathcal{K}$, a set of labels, for each subject $\mathcal{C}$.\vfill
\State Initialization: $stack=\emptyset, D=\emptyset$;
\For {$c\in \mathcal{C}$}
    \State Collect data on $c$: $D_c \longleftarrow sensor(c)$;
    \State Split dataset: $D_c=D^{train}_c \vee D^{test}_c$;
    \For {$m \in \mathcal{M}$}
        \State ${m^{train}_c\longleftarrow D^{train}_c }$;
        \State ${m^{test}_c \longleftarrow D^{test}_c }$;
        \State { $c^{\mathcal{K}} \longleftarrow  f(c)$};
        \State ${stack=stack\cup\{c^{\mathcal{K}}\}}$;
    \EndFor
\EndFor

\State{$homogeneous\_stack = stack({\{c^{\mathcal{K}}_i\}},{\{c^{\mathcal{K}}_i\}},{\{c^{\mathcal{K}}_i\}}), i\in\{1,2,\dots,|\mathcal{M}|\}$};

\For {$m \in \mathcal{M}$}
    \State ${m^{train}_g \longleftarrow homogeneous\_stack }$;
    \State ${m^{test}_g \longleftarrow D_{unseen\_c} }$;
    \State {$unseen\_c^{\mathcal{K}} \longleftarrow  f(unseen\_c)$};
\EndFor

\State{$heterogeneous\_stack = stack({\{c^{\mathcal{K}}_i\}},{\{c^{\mathcal{K}}_j\}},{\{c^{\mathcal{K}}_k\}}),\ \  {i,j,k}\in\{1,2,\dots,|\mathcal{M}|\}$};
\For {$m \in \mathcal{M}$}
    \State ${m^{train}_g \longleftarrow heterogeneous\_stack}$;
    \State ${m^{test}_g \longleftarrow D_{unseen\_c} }$;
    \State {$unseen\_c^{\mathcal{K}} \longleftarrow  f(unseen\_c)$};
\EndFor
\State {$Return$} $stack, unseen_c^K$; 
\end{algorithmic}
\end{algorithm}

The Adam adaptive optimizer was used for all three models implemented. The optimizer ensembles AdaGrad and RMSProp optimizers were implemented to deal with sparse data. Each of the three AI models was trained locally in each iteration and their performances were evaluated. These local models' weights were then aggregated based on their accuracy and forwarded to build a global model. Finally, the global model was trained on unseen data and evaluated, and is the final step of the federated architecture design shown in Fig.~\ref{fig:federated_architecture}.

In this study, the proposed FedStack Framework adopted the above-discussed AI models to train clients’ data and evaluate their performance. Instead of passing the client’s data, the local model's predictions were passed to the global model for training. With this strategy, client data will not leave their device and so will protect their privacy. Clients can train models to their requirements and pass the predicted results to the global model. This enables the personalization of client data modeling. Unlike the FedAVG concept where the same architectural models are trained across global and local models, FedStack supports heterogeneous architectures across local and global models. This will not affect federated learning, as the models are trained with predictions. Hence, the three AI models were used for both local and global models.

\subsubsection{FedStack Algorithm}

The adopted FedStack framework for personalized patient monitoring was achieved using Algorithm~\ref{alg:cap}. The algorithm presented local and global AI model execution with input client data with a set of labels. Line 1 to 11 shows the iteration of clients' data for AI modeling and stores each model's predictions. Line 12 presents the homogeneous (identical architectural models) stacking, where ${c^{\mathcal{K}}_i}$ denotes the prediction of a local model derived from Line 8 and $i$ in all the model predictions denotes identical architectural models. Line 13 to 17 use the homogeneously stacked models prediction to train global model $m_g$ to classify unseen client data ${\{unseen\_c^{\mathcal{K}}}\}$ and test its performance. Line 18 presents the heterogeneous (non-identical architectural models) stacking, where ${{\{c^{\mathcal{K}}_i\}},{\{c^{\mathcal{K}}_j\}},{\{c^{\mathcal{K}}_k\}}}$ denotes the predictions of a local models derived from Line 8 and $i,j,k$ in the model predictions denotes non-identical architectural models. Line 19 to 23 use the heterogeneously stacked models prediction to train global model $m_g$ to classify unseen client data ${\{unseen\_c^{\mathcal{K}}}\}$ and test its performance. Lines 13 to 17 and 19 to 23 present the global model training and evaluation with homogeneously stacked and heterogeneously stacked predictions, respectively. 

\section{Experimental Design}\label{exp-design}



\begin{table}[!ht]
\centering
\caption{Subject 1 Dataset—Top 5 values.}
\label{tab:dataset}
\scriptsize
\resizebox{\textwidth}{!}{
\begin{tabular}{@{}cccccc@{}}
\toprule
\textbf{Attributes} &
  \multicolumn{5}{c}{\textbf{Top 5 values}} \\ \midrule
\multicolumn{1}{c}{\textbf{C\_Sen\_AX}} &
  \multicolumn{1}{c}{{\color[HTML]{212121} -9.8184}} &
  \multicolumn{1}{c}{{\color[HTML]{212121} -9.8489}} &
  \multicolumn{1}{c}{{\color[HTML]{212121} -9.6602}} &
  \multicolumn{1}{c}{{\color[HTML]{212121} -9.6507}} &
  \multicolumn{1}{c}{{\color[HTML]{212121} -9.7030}} \\ \midrule
\multicolumn{1}{c}{\textbf{C\_Sen\_AY}} &
  \multicolumn{1}{c}{{\color[HTML]{212121} 0.009971}} &
  \multicolumn{1}{c}{{\color[HTML]{212121} 0.524040}} &
  \multicolumn{1}{c}{{\color[HTML]{212121} 0.181850}} &
  \multicolumn{1}{c}{{\color[HTML]{212121} 0.214220}} &
  \multicolumn{1}{c}{{\color[HTML]{212121} 0.303890}} \\ \midrule
\multicolumn{1}{c}{\textbf{C\_Sen\_AZ}} &
  \multicolumn{1}{c}{{\color[HTML]{212121} 0.29563}} &
  \multicolumn{1}{c}{{\color[HTML]{212121} 0.37348}} &
  \multicolumn{1}{c}{{\color[HTML]{212121} 0.43742}} &
  \multicolumn{1}{c}{{\color[HTML]{212121} 0.24033}} &
  \multicolumn{1}{c}{{\color[HTML]{212121} 0.31156}} \\ \midrule
\multicolumn{1}{c}{\textbf{LA\_Sen\_AX}} &
  \multicolumn{1}{c}{{\color[HTML]{212121} 2.1849}} &
  \multicolumn{1}{c}{{\color[HTML]{212121} 2.3876}} &
  \multicolumn{1}{c}{{\color[HTML]{212121} 2.4086}} &
  \multicolumn{1}{c}{{\color[HTML]{212121} 2.1814}} &
  \multicolumn{1}{c}{{\color[HTML]{212121} 2.4173}} \\ \midrule
\multicolumn{1}{c}{\textbf{LA\_Sen\_AY}} &
  \multicolumn{1}{c}{{\color[HTML]{212121} -9.6967}} &
  \multicolumn{1}{c}{{\color[HTML]{212121} -9.5080}} &
  \multicolumn{1}{c}{{\color[HTML]{212121} -9.5674}} &
  \multicolumn{1}{c}{{\color[HTML]{212121} -9.4301}} &
  \multicolumn{1}{c}{{\color[HTML]{212121} -9.3889}} \\ \midrule
\multicolumn{1}{c}{\textbf{LA\_Sen\_AZ}} &
  \multicolumn{1}{c}{{\color[HTML]{212121} 0.63077}} &
  \multicolumn{1}{c}{{\color[HTML]{212121} 0.68389}} &
  \multicolumn{1}{c}{{\color[HTML]{212121} 0.68113}} &
  \multicolumn{1}{c}{{\color[HTML]{212121} 0.55031}} &
  \multicolumn{1}{c}{{\color[HTML]{212121} 0.71098}} \\ \midrule
\multicolumn{1}{c}{\textbf{LA\_Sen\_GX}} &
  \multicolumn{1}{c}{{\color[HTML]{212121} 0.103900}} &
  \multicolumn{1}{c}{{\color[HTML]{212121} 0.085343}} &
  \multicolumn{1}{c}{{\color[HTML]{212121} 0.085343}} &
  \multicolumn{1}{c}{{\color[HTML]{212121} 0.085343}} &
  \multicolumn{1}{c}{{\color[HTML]{212121} 0.085343}} \\ \midrule
\multicolumn{1}{c}{\textbf{LA\_Sen\_GY}} &
  \multicolumn{1}{c}{{\color[HTML]{212121} -0.84053}} &
  \multicolumn{1}{c}{{\color[HTML]{212121} -0.83865}} &
  \multicolumn{1}{c}{{\color[HTML]{212121} -0.83865}} &
  \multicolumn{1}{c}{{\color[HTML]{212121} -0.83865}} &
  \multicolumn{1}{c}{{\color[HTML]{212121} -0.83865}} \\ \midrule
\multicolumn{1}{c}{\textbf{LA\_Sen\_GZ}} &
  \multicolumn{1}{c}{{\color[HTML]{212121} -0.68762}} &
  \multicolumn{1}{c}{{\color[HTML]{212121} -0.68369}} &
  \multicolumn{1}{c}{{\color[HTML]{212121} -0.68369}} &
  \multicolumn{1}{c}{{\color[HTML]{212121} -0.68369}} &
  \multicolumn{1}{c}{{\color[HTML]{212121} -0.68369}} \\ \midrule
\multicolumn{1}{c}{\textbf{LA\_Sen\_MY}} &
  \multicolumn{1}{c}{{\color[HTML]{212121} -0.370000}} &
  \multicolumn{1}{c}{{\color[HTML]{212121} -0.197990}} &
  \multicolumn{1}{c}{{\color[HTML]{212121} -0.374170}} &
  \multicolumn{1}{c}{{\color[HTML]{212121} -0.017271}} &
  \multicolumn{1}{c}{{\color[HTML]{212121} -0.374390}} \\ \midrule
\multicolumn{1}{c}{\textbf{LA\_Sen\_MY.1}} &
  \multicolumn{1}{c}{{\color[HTML]{212121} -0.36327}} &
  \multicolumn{1}{c}{{\color[HTML]{212121} -0.18151}} &
  \multicolumn{1}{c}{{\color[HTML]{212121} 0.18723}} &
  \multicolumn{1}{c}{{\color[HTML]{212121} 0.18366}} &
  \multicolumn{1}{c}{{\color[HTML]{212121} -0.54671}} \\ \midrule
\multicolumn{1}{c}{\textbf{LA\_Sen\_MZ}} &
  \multicolumn{1}{c}{{\color[HTML]{212121} 0.29963}} &
  \multicolumn{1}{c}{{\color[HTML]{212121} 0.58298}} &
  \multicolumn{1}{c}{{\color[HTML]{212121} 0.43851}} &
  \multicolumn{1}{c}{{\color[HTML]{212121} 0.57571}} &
  \multicolumn{1}{c}{{\color[HTML]{212121} 0.44586}} \\ \midrule
\multicolumn{1}{c}{\textbf{RLA\_Sen\_AX}} &
  \multicolumn{1}{c}{{\color[HTML]{212121} -8.6499}} &
  \multicolumn{1}{c}{{\color[HTML]{212121} -8.6275}} &
  \multicolumn{1}{c}{{\color[HTML]{212121} -8.5055}} &
  \multicolumn{1}{c}{{\color[HTML]{212121} -8.6279}} &
  \multicolumn{1}{c}{{\color[HTML]{212121} -8.7008}} \\ \midrule
\multicolumn{1}{c}{\textbf{RLA\_Sen\_AY}} &
  \multicolumn{1}{c}{{\color[HTML]{212121} -4.5781}} &
  \multicolumn{1}{c}{{\color[HTML]{212121} -4.3198}} &
  \multicolumn{1}{c}{{\color[HTML]{212121} -4.2772}} &
  \multicolumn{1}{c}{{\color[HTML]{212121} -4.3163}} &
  \multicolumn{1}{c}{{\color[HTML]{212121} -4.1459}} \\ \midrule
\multicolumn{1}{c}{\textbf{RLA\_Sen\_AZ}} &
  \multicolumn{1}{c}{{\color[HTML]{212121} 0.187760}} &
  \multicolumn{1}{c}{{\color[HTML]{212121} 0.023595}} &
  \multicolumn{1}{c}{{\color[HTML]{212121} 0.275720}} &
  \multicolumn{1}{c}{{\color[HTML]{212121} 0.367520}} &
  \multicolumn{1}{c}{{\color[HTML]{212121} 0.407290}} \\ \midrule
\multicolumn{1}{c}{\textbf{RLA\_Sen\_GX}} &
  \multicolumn{1}{c}{{\color[HTML]{212121} -0.44902}} &
  \multicolumn{1}{c}{{\color[HTML]{212121} -0.44902}} &
  \multicolumn{1}{c}{{\color[HTML]{212121} -0.44902}} &
  \multicolumn{1}{c}{{\color[HTML]{212121} -0.45686}} &
  \multicolumn{1}{c}{{\color[HTML]{212121} -0.45686}} \\ \midrule
\multicolumn{1}{c}{\textbf{RLA\_Sen\_GY}} &
  \multicolumn{1}{c}{{\color[HTML]{212121} -1.0103}} &
  \multicolumn{1}{c}{{\color[HTML]{212121} -1.0103}} &
  \multicolumn{1}{c}{{\color[HTML]{212121} -1.0103}} &
  \multicolumn{1}{c}{{\color[HTML]{212121} -1.0082}} &
  \multicolumn{1}{c}{{\color[HTML]{212121} -1.0082}} \\ \midrule
\multicolumn{1}{c}{\textbf{RLA\_Sen\_GZ}} &
  \multicolumn{1}{c}{{\color[HTML]{212121} 0.034483}} &
  \multicolumn{1}{c}{{\color[HTML]{212121} 0.034483}} &
  \multicolumn{1}{c}{{\color[HTML]{212121} 0.034483}} &
  \multicolumn{1}{c}{{\color[HTML]{212121} 0.025862}} &
  \multicolumn{1}{c}{{\color[HTML]{212121} 0.025862}} \\ \midrule
\multicolumn{1}{c}{\textbf{RLA\_Sen\_MY}} &
  \multicolumn{1}{c}{{\color[HTML]{212121} -2.35000}} &
  \multicolumn{1}{c}{{\color[HTML]{212121} -2.16320}} &
  \multicolumn{1}{c}{{\color[HTML]{212121} -1.61750}} &
  \multicolumn{1}{c}{{\color[HTML]{212121} -1.07710}} &
  \multicolumn{1}{c}{{\color[HTML]{212121} -0.53684}} \\ \midrule
\multicolumn{1}{c}{\textbf{RLA\_Sen\_MY.1}} &
  \multicolumn{1}{c}{{\color[HTML]{212121} -1.610200}} &
  \multicolumn{1}{c}{{\color[HTML]{212121} -0.882540}} &
  \multicolumn{1}{c}{{\color[HTML]{212121} -0.165620}} &
  \multicolumn{1}{c}{{\color[HTML]{212121} 0.006945}} &
  \multicolumn{1}{c}{{\color[HTML]{212121} 0.175900}} \\ \midrule
\multicolumn{1}{c}{\textbf{RLA\_Sen\_MZ}} &
  \multicolumn{1}{c}{{\color[HTML]{212121} -0.030899}} &
  \multicolumn{1}{c}{{\color[HTML]{212121} 0.326570}} &
  \multicolumn{1}{c}{{\color[HTML]{212121} -0.030693}} &
  \multicolumn{1}{c}{{\color[HTML]{212121} -0.382620}} &
  \multicolumn{1}{c}{{\color[HTML]{212121} -1.095500}} \\ \midrule
\textbf{Label} &
  {\color[HTML]{212121} 0} &
  {\color[HTML]{212121} 0} &
  {\color[HTML]{212121} 0} &
  {\color[HTML]{212121} 0} &
  {\color[HTML]{212121} 0} \\ \bottomrule
\end{tabular}
}

\end{table}
\begin{table}[!ht]\centering
\caption{Label Activities}\label{tab:Labels}
\scriptsize
\resizebox{60mm}{!}{
\begin{tabular}{lrr}\toprule
Standing still &act-1 \\
Sitting and relaxing &act-2 \\
Lying down &act-3 \\
Walking &act-4 \\
Climbing stairs &act-5 \\
Waist bends forward &act-6 \\
Frontal elevation of arms &act-7 \\
Knees bending (crouching) &act-8 \\
Cycling &act-9 \\
Jogging &act-10 \\
Running &act-11 \\
Jump front \& back &act-12 \\
\bottomrule
\end{tabular}
}
\end{table}
\subsection{Dataset}
This study was conducted on MHEALTH (Mobile HEALTH) dataset, a benchmark dataset on human behavior analysis with multi-modal sensors~\cite{banos2014mhealthdroid}\cite{banos2015design}. It is fashioned upon the Banos et al.\cite{banos2015design} studies where data were collected on ten different subjects while performing natural activities with three sensors placed on the subject's chest, right wrist, and left ankle. In addition to the physical activities, vital signs were recorded with 2-lead electrocardiogram (ECG) measurements using the sensor placed in the chest area. However, the ECG measurements were not an aim of this research and as such two attributes of 2-lead ECG signal data were excluded, thus the number of attributes was reduced to 21 and included a label as shown in Tab.~\ref{tab:dataset}. The sensor placed in the chest area has three attributes comprising tri-axial data of acceleration (x-axis, y-axis, z-axis). Similarly, the sensors at the left ankle and right wrist have motion attributes of acceleration (x-axis, y-axis, z-axis), gyro (x-axis, y-axis, z-axis), and magnetometer (x-axis, y-axis, z-axis). Based on these tri-axial attributes, the authors labelled 12 natural activities like standing still, lying down, walking, and climbing stairs using motion like acceleration, rate of turn, and magnetic field orientation experienced by diverse body parts as shown in Tab.~\ref{tab:Labels}. Each of these labels was numbered from 0 to 12. The dataset generalized all daily activities to cover a wide range of body parts in each activity, speed, and intensity of actions. 

The main aim of designing this RPM system is to monitor multiple patients' physical activities and vital signs in an acute mental health facility. Therefore, the dataset comprises most of the physical activities typical of routine day-to-day life. The study offers an alternative method for classifying human body motion to previous research methods. There were no constraints on the data except the subject's effort while performing the activities, and all sessions were video recorded. Records were assigned a label 0 for those with null activities to differentiate them from other activities in the dataset. 

\subsection{Data Preparation}
The benchmark dataset comprises one dataset for each subject, which sums up the count of datasets to ten log files. To ease the implementation process, the log files were transformed into CSV files using python code and read the CSV files as a data frame for each subject dataset. All the records with null activities were excluded based on the label value 0. To ensure the consistency in values of independent variables, all variables were standardized using StandardScaler\footnote{https://scikit-learn.org/stable/modules/generated/sklearn.preprocessing.\\StandardScaler.html} methods in the sklearn package. 

\begin{table}[!ht]\centering
\caption{Subject 1—Dataset PCA values}\label{tab:PCA}
\scriptsize
\begin{tabular}{lrrrrrr}\toprule
\textbf{Principal} &\multicolumn{3}{c}{\textbf{}} &\textbf{Eigen} &\textbf{Cumulative} \\
\textbf{components} &\multicolumn{3}{c}{\textbf{Top 3 values}} &\textbf{values} &\textbf{eigen values} \\\midrule
PC 1 &1.549 &1.618 &1.595 &0.167 &0.167 \\
PC 2 &0.133 &0.089 &0.066 &0.122 &0.289 \\
PC 3 &-1.047 &-0.987 &-0.959 &0.107 &0.396 \\
PC 4 &-0.838 &-0.86 &-0.885 &0.092 &0.488 \\
PC 5 &-1.142 &-1.094 &-1.104 &0.082 &0.569 \\
PC 6 &0.929 &0.875 &0.848 &0.061 &0.63 \\
PC 7 &-0.664 &-0.697 &-0.722 &0.054 &0.684 \\
PC 8 &-1.057 &-1.054 &-1.045 &0.044 &0.728 \\
PC 9 &-0.276 &-0.273 &-0.288 &0.042 &0.771 \\
PC 10 &-0.128 &-0.186 &-0.255 &0.034 &0.804 \\
PC 11 &-0.004 &0.022 &0.058 &0.032 &0.836 \\
PC 12 &0.245 &0.215 &0.212 &0.029 &0.865 \\
PC 13 &-0.044 &-0.008 &-0.022 &0.026 &0.891 \\
PC 14 &-0.176 &-0.034 &0.003 &0.024 &0.915 \\
PC 15 &-0.171 &-0.206 &-0.21 &0.018 &0.934 \\
PC 16 &-0.26 &-0.296 &-0.313 &0.017 &\textbf{0.951} \\
PC 17 &-0.163 &-0.166 &-0.163 &0.016 &0.966 \\
PC 18 &-0.09 &-0.066 &-0.059 &0.014 &0.981 \\
PC 19 &-0.067 &-0.076 &-0.067 &0.008 &0.989 \\
PC 20 &0.1 &0.109 &0.124 &0.006 &0.995 \\
PC 21 &-0.008 &-0.028 &0.008 &0.005 &1 \\
\bottomrule
\end{tabular}
\end{table}

Using the PCA technique, 21 principal components were extracted from 21 features of tri-axial data from the three different sensors, as shown in Tab.~\ref{tab:PCA}. In this study, 95\% of the total variability was explained by 16 principal components. With this, the feature dimensionality was reduced and filtered noise was conducted by selecting the 16 principal components for data modeling. All the tri-axial attributes were considered as features and sparse the multi-class label variable into binary labels. The 16 principal components and the 12 binary labels were transformed into test and train data by splitting train data as 80\% and test data as 20\%. The transformed data were then fed to AI models in the data modeling step.


\subsection{Data Modeling}
In the data modeling step, three different AI models ANN, CNN, and LSTM is known for their efficient performance in HAR with strong evidence from research community works. These AI models were trained individually on each subject dataset. Out of 10 subjects in the benchmark dataset, nine subjects were considered individual clients, and one subject dataset was trained to the global model. All three AI models discussed in the framework section were trained, and their performance was evaluated on the nine client datasets. As shown in the architecture of Fig.~\ref{fig:federated_architecture}, a local model is denoted, $L_{i}$ where $i$ value ranges from 0 to 9 clients. It is built on each client with one AI model at a time and their performances compared with FL.



The ANN model was executed with three layers: an input layer, a hidden layer, and an output layer. The model used the activation function LeakyReLU to avoid the zero input values of negative attributes in the traditional ReLU function. The Adam algorithm was chosen as the optimizing method in all three AI models in this study. The second AI model adopted was the CNN model. Each axis attribute of acceleration, gyro, and magnetometer was fed to a 1-dimensional convolutional layer with linear activation and the signal passed to the LeakyReLU function with a small alpha value of 0.1. MaxPool1D was employed in the next layer to downsample the input representation and reduce the pool size to 2. The three layers were repeated with a different number of neurons in each convolutional layer. Following this, the pooled feature was flattened before it was forwarded to the output layer of the deep learning model. Recurrent Neural Network (RNN) based Bi-directional LSTM was also trained on each client dataset and the model performance was evaluated. This DL model was executed with LeakyReLU activation in the input layer. The dropout method was added in between the input and output layers with a dropout percentage of 0.5. Softmax was applied to the output layer as an activation function in all three AI models to output the probability of the label classification. The binary label classification was optimized using the threshold capacity. 


After individual client modeling, each model prediction was recorded. All the client model predictions were stacked homogeneously and heterogeneously and passed to the global model. As mentioned earlier in this subsection, one of the subjects' data was used to train the global model, which is unseen data to the global model. The model was then trained with stacked predictions of local models. In each iteration, the FL process was executed with each AI model on each of the nine clients.

Python programming language (version 3.8) was used for data preparation, dimensionality reduction, and data modeling including FL and model evaluation. TensorFlow and Keras packages were imported to execute all three AI models. Communication between local models with a global model in FL was based upon the proposed novel FedStack algorithm.

\subsection{Baseline Models}
The proposed research design was evaluated with baseline models with state-of-art performances in human activity recognition. 
\begin{itemize}
    \item Ronao et al.~\cite{ronao2016human} proposed a deep convolutional neural network (convnet) to classify 1-D sensor data into physical activities and achieved an accuracy of 94.9\% on raw sensor data, outperforming state-of-the-art techniques in HAR. The performance was slightly improved to 97.75\% with the temporal fast Fourier transform of the dataset. The authors also show that increasing the convolutional layers increases performance, but the complexity of the derived features decreases with every additional layer in their study~\cite{ronao2015deep}. 
    \item Jiang et al.~\cite{jiang2015human} proposed a novel approach to assemble signal sequences of accelerometers and gyroscopes into a novel activity image using Deep Convolutional Neural Networks (DCNN) and achieved an accuracy of 95.2\%. 
    \item Almaslukh et al.~\cite{almaslukh2017effective} proposed a stacked autoencoder (SAE) to achieve high computational cost with low computation cost and yield an accuracy of 97.5\%. 
    \item Ignatov et al.~\cite{ignatov2018real} proposed a CNN model for local feature extraction and combine them with statistical features. This approach outperformed state-of-the-art works with an accuracy of 96.1\% in activity recognition.
    \item  Anguita et al.~\cite{anguita2013public} proposed a traditional machine learning algorithm SVM to classify human activities on their sensor dataset Activities of Daily Living (ADL). The model achieved an accuracy of 96.4\% dominating previously discussed works using the deep learning CNN model. 
    \item Cho et al.~\cite{cho2018divide} proposed multiple 1-D CNN models for human activity recognition at different stages of the experiment to learn abstract activities and then learn individual activities. This design achieved an accuracy of 97.6\%.
    
\end{itemize}
All these baseline models had the best performance in human activity recognition with deep learning and traditional machine learning activities. The proposed FedStack design with a similar deep learning model was evaluated with these state-of-work performances.

\subsection{Performance Metrics}
Confusion matrix was used to evaluate the classification models. Each subject dataset label was transformed using dummy variable encoding that led to multiple binary labels for each record. To evaluate the multiple binary label classification, a multi-label confusion matrix was used. It was imported from sklearn metrics and required two inputs, actual data and predicted data. The classification models were evaluated on learned data as well as unseen data. The multi-label confusion matrix for each transformed target variable.


Based on the confusion matrix for each target variable, metrics like balanced accuracy, precision, recall, f1-score, and support were estimated using the classification\_report method from sklearn metrics~\cite{hossin2015review}. All metrics were estimated for the classification of each physical activity. Hence, the classification model performance was evaluated by classifying each activity individually.

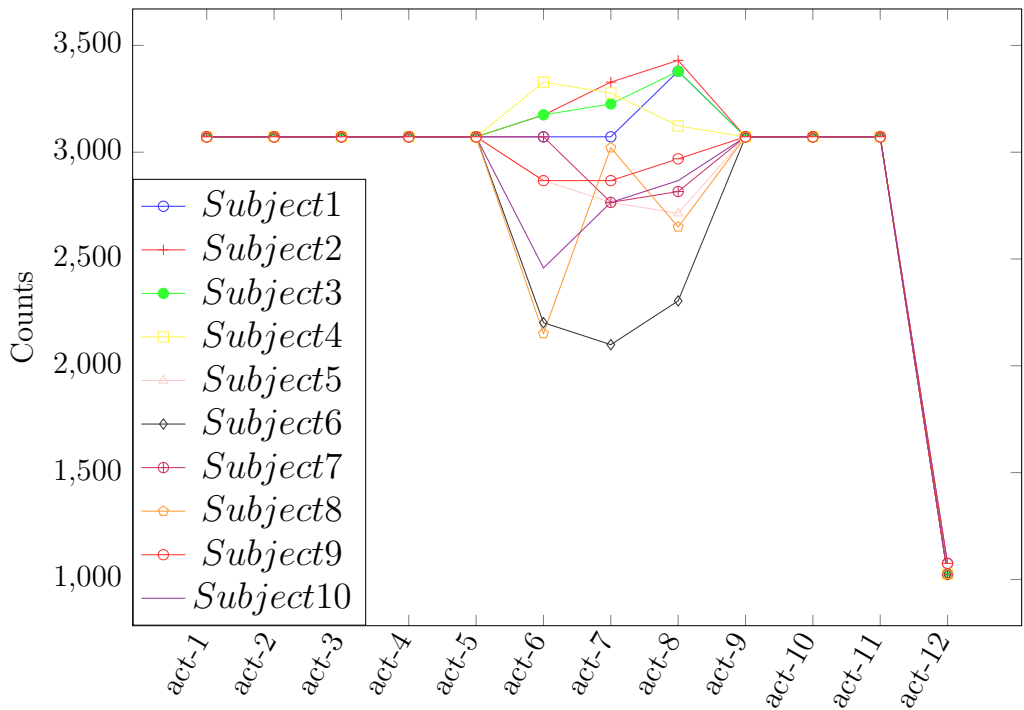
\begin{figure}[!ht]
    \centering
\pgfplotstableread[col sep=comma,]{label_dist.csv}\datatable
\resizebox{\columnwidth}{!}{%
\begin{tikzpicture}
\begin{axis}[
    width=\textwidth,
    height=10cm,
    xtick=data,
    xticklabels from table={\datatable}{Label},
    x tick label style={font=\normalsize, rotate=60, anchor=east},
    legend style={nodes={scale=1.2, transform shape},at={(0.0, 0.0)},anchor=south west},
    ylabel={Counts}]
    
    \addplot [mark=o, blue!80 ] table [x expr=\coordindex, y={Subject 1}]{\datatable};
    \addlegendentry{$Subject 1$}
    
    \addplot [mark=+, red!80 ] table [x expr=\coordindex, y={Subject 2}]{\datatable};
    \addlegendentry{$Subject 2$}
    
     \addplot [mark=*, green!80 ] table [x expr=\coordindex, y={Subject 3}]{\datatable};
    \addlegendentry{$Subject 3$}
     \addplot [mark=square, yellow!80 ] table [x expr=\coordindex, y={Subject 4}]{\datatable};
    \addlegendentry{$Subject 4$}
     \addplot [mark=triangle, pink!80 ] table [x expr=\coordindex, y={Subject 5}]{\datatable};
    \addlegendentry{$Subject 5$} 
    
    \addplot [mark=diamond, black!80 ] table [x expr=\coordindex, y={Subject 6}]{\datatable};
    \addlegendentry{$Subject 6$}
     \addplot [mark=oplus, purple!80 ] table [x expr=\coordindex, y={Subject 7}]{\datatable};
    \addlegendentry{$Subject 7$}
     \addplot [mark=pentagon, orange!80 ] table [x expr=\coordindex, y={Subject 8}]{\datatable};
    \addlegendentry{$Subject 8$}
     \addplot [mark=halfcircle, red!80 ] table [x expr=\coordindex, y={Subject 9}]{\datatable};
    \addlegendentry{$Subject 9$}
     \addplot [mark=cubes, violet!80 ] table [x expr=\coordindex, y={Subject 10}]{\datatable};
    \addlegendentry{$Subject 10$}

\end{axis}
\end{tikzpicture}
}
    \caption{Subjects—Label Distribution}
    \label{fig:label_dist}
\end{figure}
\section{Experimental Results Analysis}\label{results}
\subsection{Experimental Results}
The study was analyzed with ten different subjects’ data using AI models. Each of the subjects was considered as an individual client, and an FL process was initiated by passing nine client models (local models) parameters to the global model. The communicated parameters were weight averaged based on each local model performance on client data. The global model was configured based on the aggregated parameters from the local models. In addition to this, both local and global models were evaluated by classifying the labels based on one sensor data at a time. 

Before the data preparation step, the label distribution of each subject data was visualized by excluding null activity with zero value as shown in Fig.~\ref{fig:label_dist}. The line chart presents subject-wise label distribution. Each subject was differentiated with different colors as shown in the legend, with the x-axis referring to the labels and the y-axis referring to the count of records with labels for each subject. Except for the four labels of waist bends forward, the frontal elevation of arms, knees bending (crouching), and jump front and back, all others are class-balanced for all subjects. The labels' waist bends forward, the frontal elevation of arms, and knees bending (crouching) look imbalanced with minor differences in numbers. Jump front and back labels had significantly reduced to about 1000 records compared to other labels.

\begin{table}[!ht]\centering
\caption{Proposed AI models performance (accuracy~\%).}\label{tab:DL_performance}

\begin{tabular}{@{}lrrr@{}}
\toprule
\rowcolor[HTML]{FFFFFF} 
\textbf{Clients} &
  \multicolumn{1}{l}{\cellcolor[HTML]{FFFFFF}\textbf{ANN}} &
  \multicolumn{1}{l}{\cellcolor[HTML]{FFFFFF}\textbf{CNN}} &
  \multicolumn{1}{l}{\cellcolor[HTML]{FFFFFF}\textbf{Bi-LSTM}} \\ \midrule
\rowcolor[HTML]{FFFFFF} 
Client 1                                                                      & 0.976          & \textbf{0.988} & 0.934 \\ \midrule
\rowcolor[HTML]{FFFFFF} 
Client 2                                                                      & 0.939          & \textbf{0.957} & 0.837 \\ \midrule
\rowcolor[HTML]{FFFFFF} 
Client 3                                                                      & 0.98           & \textbf{0.995} & 0.946 \\ \midrule
\rowcolor[HTML]{FFFFFF} 
Client 4                                                                      & 0.991          & \textbf{0.997} & 0.948 \\ \midrule
\rowcolor[HTML]{FFFFFF} 
Client 5                                                                      & 0.966          & \textbf{0.989} & 0.897 \\ \midrule
\rowcolor[HTML]{FFFFFF} 
Client 6                                                                      & \textbf{0.984} & \textbf{0.984} & 0.928 \\ \midrule
Client 7                                                                      & \textbf{0.998} & \textbf{0.998} & 0.986 \\ \midrule
Client 8                                                                      & 0.985          & \textbf{0.991} & 0.951 \\ \midrule
Client 9                                                                      & 0.994          & \textbf{0.995} & 0.96  \\ \midrule
\begin{tabular}[c]{@{}l@{}}Homogeneous Stacked \\ Global Model\end{tabular}   & 0.967          & \textbf{0.976} & 0.909 \\ \midrule
\begin{tabular}[c]{@{}l@{}}Heterogeneous Stacked \\ Global Model\end{tabular} & \textbf{0.996} & \textbf{0.996} & 0.986 \\ \bottomrule
\end{tabular}
\end{table}


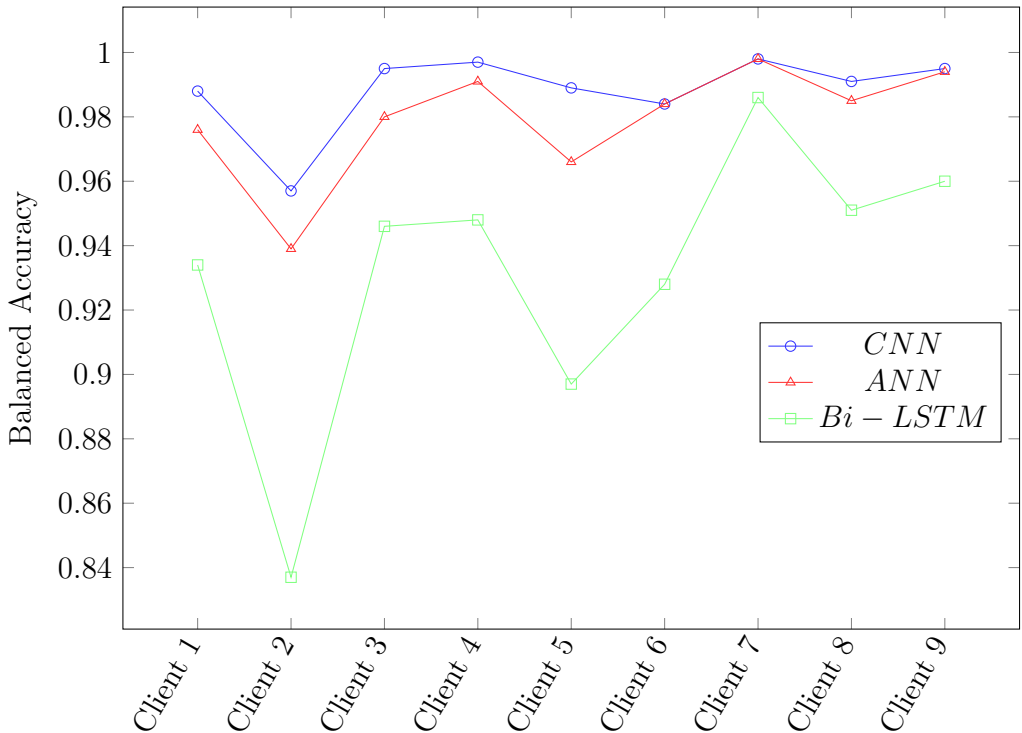
\begin{figure}[!ht]
    \centering
\pgfplotstableread[col sep=comma,]{local_global_model.csv}\datatable
\resizebox{\columnwidth}{!}{%
\begin{tikzpicture}
\begin{axis}[
    width=\textwidth,
    height=10cm,
    xtick=data,
    xticklabels from table={\datatable}{Clients},
    x tick label style={font=\normalsize, rotate=60, anchor=east},
    legend style={at={(0.98,0.3)},anchor=south east},
    ylabel={Balanced Accuracy}]
    
    \addplot [mark=o, blue!80 ] table [x expr=\coordindex, y={CNN}]{\datatable};
    \addlegendentry{$CNN$}
    
    \addplot [mark=triangle, red!80] table [x expr=\coordindex, y={ANN}]{\datatable};
    \addlegendentry{$ANN$}
    
    \addplot [mark=square, green!50 ] table [x expr=\coordindex, y={Bi-LSTM}]{\datatable};
    \addlegendentry{$Bi-LSTM$}
\end{axis}
\end{tikzpicture}
}
\caption{Local and global model accuracies obtained for various clients.}
\label{fig:local_global_model}
\end{figure}

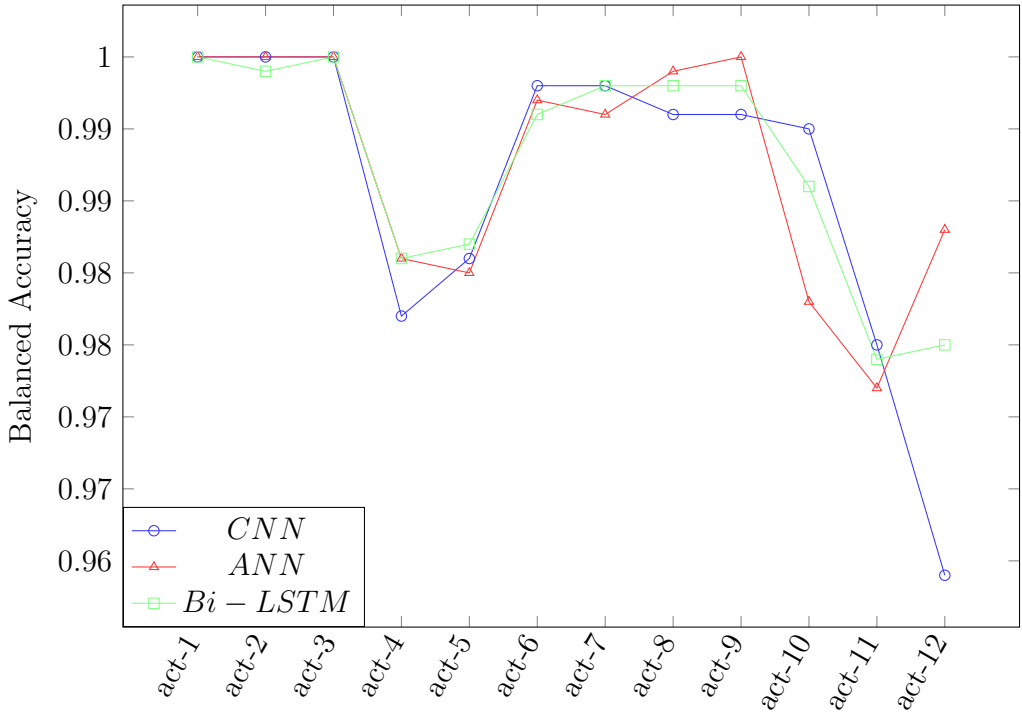
\begin{figure}[!ht]
\centering
\pgfplotstableread[col sep=comma,]{HomoStack_Global.csv}\datatable
\resizebox{\columnwidth}{!}{%
\begin{tikzpicture}
\begin{axis}[
width=\textwidth,
height=10cm,
xtick=data,
xticklabels from table={\datatable}{Label},
x tick label style={font=\normalsize, rotate=60, anchor=east},
legend style={nodes={scale=1.0, transform shape},at={(0.0, 0.0)},anchor=south west},
ylabel={Balanced Accuracy}]

\addplot [mark=o, blue!80 ] table [x expr=\coordindex, y={CNN}]{\datatable};
\addlegendentry{$CNN$}

\addplot [mark=triangle, red!80] table [x expr=\coordindex, y={NN}]{\datatable};
\addlegendentry{$ANN$}

\addplot [mark=square, green!50 ] table [x expr=\coordindex, y={Bi-LSTM}]{\datatable};
\addlegendentry{$Bi-LSTM$}
\end{axis}
\end{tikzpicture}
}
\caption{Homogeneous stacked global model}
\label{fig:Homo_stacked_global_model}
\end{figure}

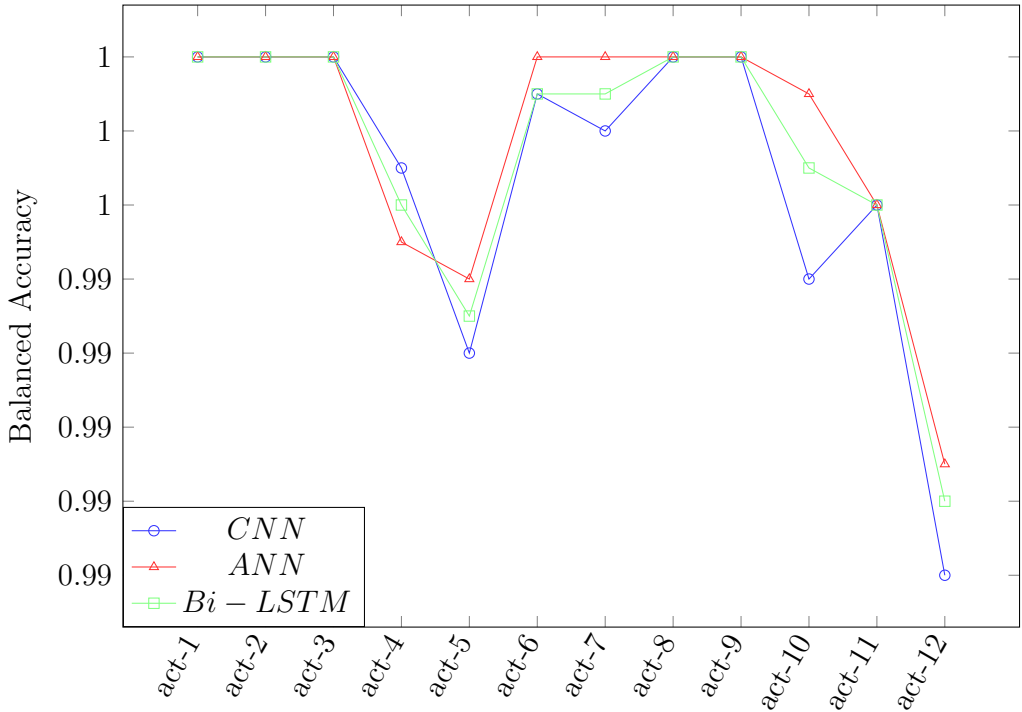
\begin{figure}[!ht]
\centering
\pgfplotstableread[col sep=comma,]{HeteroStack_Global.csv}\datatable
\resizebox{\columnwidth}{!}{%
\begin{tikzpicture}
\begin{axis}[
width=\textwidth,
height=10cm,
xtick=data,
xticklabels from table={\datatable}{Label},
x tick label style={font=\normalsize, rotate=60, anchor=east},
legend style={nodes={scale=1.0, transform shape},at={(0.0, 0.0)},anchor=south west},
ylabel={Balanced Accuracy}]

\addplot [mark=o, blue!80 ] table [x expr=\coordindex, y={CNN}]{\datatable};
\addlegendentry{$CNN$}

\addplot [mark=triangle, red!80] table [x expr=\coordindex, y={NN}]{\datatable};
\addlegendentry{$ANN$}

\addplot [mark=square, green!50 ] table [x expr=\coordindex, y={Bi-LSTM}]{\datatable};
\addlegendentry{$Bi-LSTM$}
\end{axis}
\end{tikzpicture}
}
\caption{Heterogeneous stacked global model}
\label{fig:Hetero_stacked_global_model}
\end{figure}

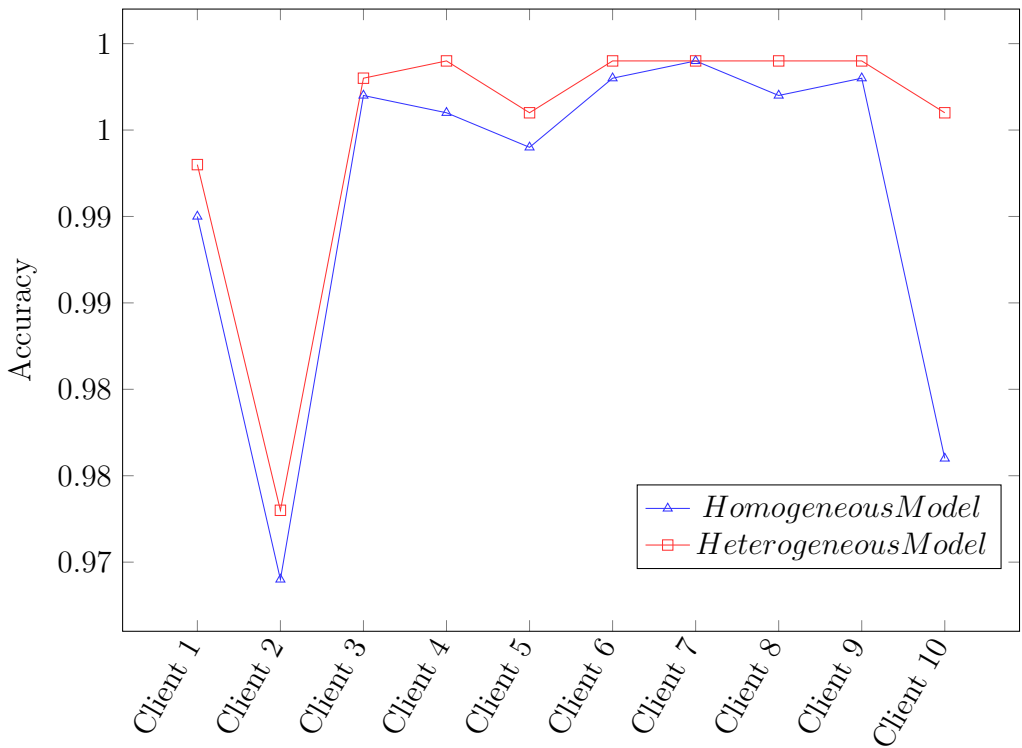
\begin{figure}[!ht]
    \centering
\pgfplotstableread[col sep=comma,]{global_cnn.csv}\datatable
\resizebox{\columnwidth}{!}{%
\begin{tikzpicture}
\begin{axis}[
    width=\textwidth,
    height=10cm,
    xtick=data,
    xticklabels from table={\datatable}{Clients},
    x tick label style={font=\normalsize, rotate=60, anchor=east},
    legend style={at={(0.98,0.1)},anchor=south east},
    ylabel={Accuracy}]
    
    \addplot [mark=triangle, blue!80 ] table [x expr=\coordindex, y={Homogeneous CNN Model}]{\datatable};
    \addlegendentry{$Homogeneous Model$}
    
    \addplot [mark=square, red!80] table [x expr=\coordindex, y={Heterogeneous CNN Model}]{\datatable};
    \addlegendentry{$Heterogeneous Model$}

\end{axis}
\end{tikzpicture}
}
\caption{Global CNN model performance.}
\label{fig:global_cnn}
\end{figure}

\begin{figure}[!htp]
     \centering
     \begin{subfigure}[b]{\textwidth}
        \centering
        \pgfplotstableread[col sep=comma,]{Chest_Sensor_Result.csv}\datatable
        \resizebox{\columnwidth}{!}{%
        \begin{tikzpicture}
        \begin{axis}[
            width=\textwidth,
            height=6cm,
            xtick=data,
            xticklabels from table={\datatable}{Label},
            x tick label style={font=\scriptsize, rotate=60, anchor=east},
            y tick label style={font=\scriptsize},
            legend style={nodes={scale=0.6, transform shape},at={(0.0, 0.0)},anchor=south west},
            ylabel={Performance Metrics}]
            
            \addplot [mark=square, blue!80 ] table [x expr=\coordindex, y={Balanced Accuracy}]{\datatable};
            \addlegendentry{$bal\_accuracy$}
            
            \addplot [mark=o, green!80 ] table [x expr=\coordindex, y={precision}]{\datatable};
            \addlegendentry{$precision$}
            
            \addplot [mark=+, red!80] table [x expr=\coordindex, y={recall}]{\datatable};
            \addlegendentry{$recall$}
            
            \addplot [mark=asterisk, black!50 ] table [x expr=\coordindex, y={f1-score}]{\datatable};
            \addlegendentry{$f1-score$}
        \end{axis}
        \end{tikzpicture}
        }
         \caption{Chest sensor}
         \label{fig:Chest_Sensor_Result}
     \end{subfigure}%
     \vfill
     \begin{subfigure}[b]{\textwidth}
         \centering
         \pgfplotstableread[col sep=comma,]{Left_Ankle_Result.csv}\datatable
        \resizebox{\columnwidth}{!}{%
        \begin{tikzpicture}
        \begin{axis}[
            width=\textwidth,
            height=6cm,
            xtick=data,
            xticklabels from table={\datatable}{Label},
            x tick label style={font=\scriptsize, rotate=60, anchor=east},
            y tick label style={font=\scriptsize},
            legend style={nodes={scale=0.6, transform shape},at={(0.18,0.0)},anchor=south},
            ylabel={Performance Metrics}]
            
            \addplot [mark=square, blue!80 ] table [x expr=\coordindex, y={Balanced Accuracy}]{\datatable};
            \addlegendentry{$bal\_accuracy$}
            
            \addplot [mark=o, green!80 ] table [x expr=\coordindex, y={precision}]{\datatable};
            \addlegendentry{$precision$}
            
            \addplot [mark=+, red!80] table [x expr=\coordindex, y={recall}]{\datatable};
            \addlegendentry{$recall$}
            
            \addplot [mark=asterisk, black!50 ] table [x expr=\coordindex, y={f1-score}]{\datatable};
            \addlegendentry{$f1-score$}
        \end{axis}
        \end{tikzpicture}
        }
         \caption{Left ankle sensor}
         \label{fig:Left_Ankle_Result}
     \end{subfigure}
     \vfill
     \begin{subfigure}[b]{\textwidth}
        
         \centering
         \pgfplotstableread[col sep=comma,]{Right_Wrist_Result.csv}\datatable
        \resizebox{\columnwidth}{!}{%
        \begin{tikzpicture}
        \begin{axis}[
            width=\textwidth,
            height=6cm,
            xtick=data,
            xticklabels from table={\datatable}{Label},
            x tick label style={font=\scriptsize, rotate=60, anchor=east},
            y tick label style={font=\scriptsize},
            legend style={nodes={scale=0.6, transform shape},at={(0.35,0.0)},anchor=south east},
            ylabel={Performance Metrics}]
            
            \addplot [mark=square, blue!80 ] table [x expr=\coordindex, y={Balanced Accuracy}]{\datatable};
            \addlegendentry{$bal\_accuracy$}
            
            \addplot [mark=o, green!80 ] table [x expr=\coordindex, y={precision}]{\datatable};
            \addlegendentry{$precision$}
            
            \addplot [mark=+, red!80] table [x expr=\coordindex, y={recall}]{\datatable};
            \addlegendentry{$recall$}
            
            \addplot [mark=asterisk, black!50 ] table [x expr=\coordindex, y={f1-score}]{\datatable};
            \addlegendentry{$f1-score$}
        \end{axis}
        \end{tikzpicture}
        }
         \caption{Right wrist sensor}
         \label{fig:Right_Wrist_Result}
     \end{subfigure}
        \caption{CNN performance at the sensor level.}
        \label{fig:CNN_Performance}
\end{figure}
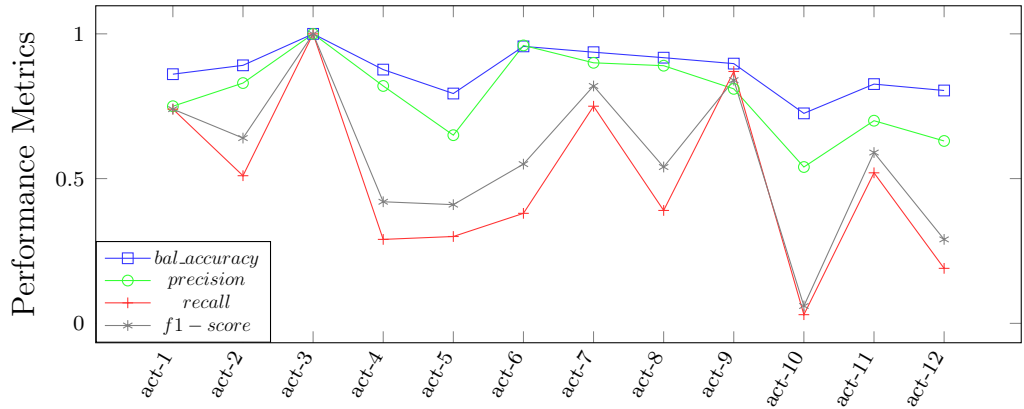
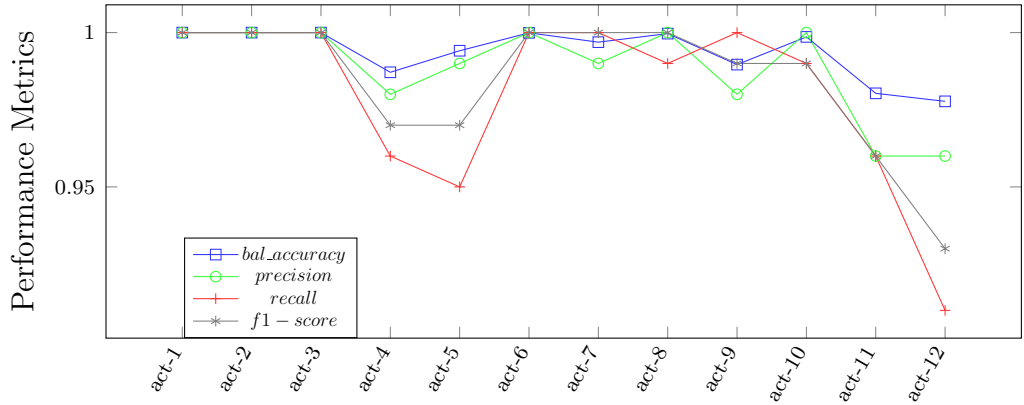
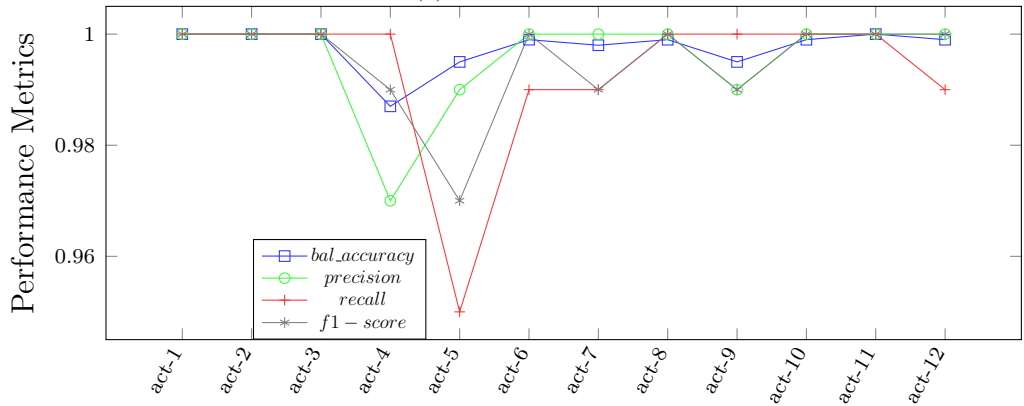
\begin{table}[!ht]\centering
\caption{Comparison of the proposed model with state-of-the-art techniques.}\label{tab:baseline}
\scriptsize
\begin{adjustbox}{max width=\columnwidth}
\begin{tabular}{lrrr}\toprule
\textbf{Reference} &\textbf{Models} &\textbf{Accuracy (\%)} \\\midrule
Ronao and Cho, 2016~\cite{ronao2016human} &3 * Conv + Dense layer &0.948 \\
Jiang and Yin, 2015~\cite{jiang2015human} &2 Conv + Dense Layer &0.952 \\
Ronao and Cho, 2015~\cite{ronao2015deep} &3 * Conv + Dense Layer &0.948 \\
Almaslukh, 2017~\cite{almaslukh2017effective} &2 * Dense Layer (SAE) &0.975 \\
Ignatov, 2018~\cite{ignatov2018real} &1 * Conv + Dense Layer &0.961 \\
Anguita et al., 2013~\cite{anguita2013public} &SVM &0.964 \\
Cho and Yoon, 2018 ~\cite{cho2018divide} &DT + 2 * CNN &0.976 \\\midrule
Proposed CNN—Client  &3 * Conv + Dense &\textbf{0.988} \\
Proposed ANN—Client  &1 * NN + Dense &\textbf{0.979}\\
Proposed Bi-LSTM—Client  &1 * Bi-LSTM + Dense &0.932\\
\begin{tabular}[c]{@{}l@{}}Proposed Homogeneous \\ Stacked CNN Global Model \end{tabular}&3 * Conv + Dense &\textbf{0.976}\\
\begin{tabular}[c]{@{}l@{}}Proposed Homogeneous \\ Stacked ANN Global Model \end{tabular}&1 * NN + Dense &0.967\\
\begin{tabular}[c]{@{}l@{}}Proposed Homogeneous \\ Stacked Bi-LSTM Global Model \end{tabular}&1 * Bi-LSTM + Dense&0.909 \\
\begin{tabular}[c]{@{}l@{}}Proposed Heterogeneous \\ Stacked CNN Global Model \end{tabular}&3 * Conv + Dense &\textbf{0.996}\\
\begin{tabular}[c]{@{}l@{}}Proposed Heterogeneous \\ Stacked ANN Global Model \end{tabular}&1 * NN + Dense&\textbf{0.996}\\
\begin{tabular}[c]{@{}l@{}}Proposed Heterogeneous \\ Stacked Bi-LSTM Global Model \end{tabular}&1 * Bi-LSTM + Dense &\textbf{0.986} \\

\bottomrule
\end{tabular}
\end{adjustbox}
\end{table}
As part of the data modeling step, three AI models ANN, CNN, and LSTM models were chosen for training and evaluation because of their robust and efficient performances in the classification of human physical activities. Each of these models was trained with individual client data, considering them as local models in FL. The model performances in terms of balanced accuracy are presented in Tab.~\ref{tab:DL_performance} and compared. The CNN model outperformed the other two AI models. ANN performance was close to the CNN model and was even equivalent in client 6, client 7, and heterogeneous stacked global model analysis. Bidirectional-LSTM had limited performance compared to the CNN and ANN models. In addition to local models, the global model performed equally with the local client models. It was built on stacked predictions of local models and trained with the unseen data of subject 10. All model performances were visually compared from the line chart shown in Fig.~\ref{fig:local_global_model}. The chart compared all 9 clients, with the y-axis denoting balanced accuracy and the x-axis are the models. Bi-LSTM model performance significantly dropped for client 2 and client 5 data classification. The ANN model had a similar trend, with its performance dipping for client 2 and client 5.

The stacked global models were trained using subject 10 data for each AI model, and their performance was evaluated as shown in Fig.~\ref{fig:Homo_stacked_global_model}~\&~\ref{fig:Hetero_stacked_global_model}. It demonstrates the classification performance of each AI model on each label activity that was calculated using a confusion matrix. The x-axis denotes the labels, and the y-axis is the balanced accuracy calculated from the confusion matrix.

The process of FL was iterated on the homogeneous and heterogeneous stacked CNN models with the best performance for all 10 clients. Local models were trained using a leave-one-out strategy where one client was left out for global model training and each of the remaining nine clients was trained to local models individually. The global models built on the stacked predictions of local models classified physical activities and achieved similar results for all 10 clients. Although the heterogeneous global model outperformed the homogeneous model, they followed a similar trend.  The line chart presents the global CNN model's accuracy in evaluating all 10 clients.

\subsection{Sensor Level—CNN Performance}
To develop an efficient system with a single sensor, the best performed global heterogeneous CNN model as shown in Fig.~\ref{fig:global_cnn} was trained with one sensor input at a time. All evaluated performance metrics of the models are shown in Fig.~\ref{fig:CNN_Performance}. The three subplots compare the performance of the federated global CNN model on the Chest sensor in Fig.~\ref{fig:Chest_Sensor_Result}, the Left Ankle sensor in Fig.~\ref{fig:Left_Ankle_Result}, and the Right Wrist sensor in Fig.~\ref{fig:Right_Wrist_Result}. Each subplot has an x-axis with 12 labels and a y-axis with a scale to show balanced accuracy, precision, recall, and f1-score.

The CNN model has considerable balanced accuracy, with an exception in classifying jogging and climbing stairs activities of chest sensor data, as shown in Fig.~\ref{fig:Chest_Sensor_Result}. Precision metrics followed a similar trend with balanced accuracy. Recall and f1-score had similar trends, with a number of fluctuations in each label classification. The CNN model performed well with left ankle sensor data input in terms of all metrics compared to chest sensor data input, as shown in Fig.~\ref{fig:Left_Ankle_Result}. There are a few exceptions in classifying walking, running, jumping front \& back, and climbing stairs. The CNN model performance in classifying the label activities using right wrist sensor data, as shown in Fig.~\ref{fig:Right_Wrist_Result}. The model was able to achieve more than 0.98 balanced accuracies in classifying the labels except walking. Precision metric had a similar drop at the walking label. Recall and f1-score had a drop for classifying the climbing stairs activity.

\subsection{Baseline Models Comparison}
The proposed AI models' performance has been compared with the state-of-art baseline model results in human activity recognition, as shown in Tab.~\ref{tab:baseline}. It has different article references with corresponding models implemented in classifying human physical activities. The proposed local models' accuracy was presented in mean accuracy. Out of all proposed AI models, CNN models outperform baseline models, both locally and globally. ANN local model and heterogeneous global model were the best performing baseline models. Bi-LSTM global models were able to perform better, but the local model needs further optimization. 

\subsection{Discussion}

The primary contribution of this study is the heterogeneous FedStack architecture shown in Fig.~\ref{fig:federated_architecture} which could process a variety of client architectures. Original federated learning has a limitation of aggregating different architectural local models due to discrepancies in layer count mismatch. The proposed FedStack algorithm overcame this challenge and outperformed baseline models. This research will also contribute to the building of an RPM system to remotely detect patient health parameters in an acute mental health facility using passive RFID tags. One of the major challenges in achieving this goal is the protection of private patient data. This proposed machine learning model was trained locally and passed only the model predictions to the global model to prevent security breaches of private data. This architecture presents an alternative to gathering both public and private data from all subjects and merging data for model building. This FL approach has the additional advantage of randomly selecting subjects in an institution and builds robust models based on communication between local models and global models, where only the model weights are shared. The globally built model predictions or parameters can also be communicated to local models so that local models can be improved. This FL process secures individual data and improves the diversity of data. The global and local models remain in continuous learning mode by updating each other with new model weights. However, the FL approach needs to strengthen its features, being a relatively recent innovation. One limitation is that the privacy rule of an FL process can be violated by reverse engineering processes, and the research community needs to explore methodologies to ensure that the features are robust~\cite{cheng2020federated}. This is an interesting FL challenge that should be addressed in future research. 

To avoid the minor class imbalance in the label distribution, the balanced accuracy method was adopted for this study to avoid minor class imbalance in the label distribution illustrated in Fig.~\ref{fig:label_dist}. As shown in Fig.~\ref{fig:Homo_stacked_global_model}~\&~\ref{fig:Hetero_stacked_global_model}, overall classification performance using a CNN model on each client and global model demonstrated the best outcomes for label classification compared to the other AI models tested. An exception to this model performance were the results for homogeneous ANN and Bi-LSTM models that appeared to perform better at classifying activities related to walking, knee bending, cycling, and jump front and back.

Classification of labels for this dataset covered a diverse range of activities. It was based upon large body movements from physical activities that are performed either consciously or unconsciously. This is an important aspect of designing the hospital-based RPM system, as patients in an acute mental health facility are quite mobile. The approach proposed in this research through classifying physical activities using AI models outperforms traditional machine learning models discussed in the literature review~\cite{sri2021performance,bulbul2018human,asim2020context,vaizman2017recognizing}.

The ANN model also demonstrated better performance when classifying labels for climbing stairs, frontal elevation, and jump front and back. The Bi-LSTM model was proficient with classifying label activities with considerable balanced accuracy, but ANN and CNN outperformed this model in all local model performances. This indicates that RNN with memory blocks needs further model optimization to enhance performance. 

The secondary aim of this study was to understand the optimum position for sensors on the body to track day-to-day activities. The dataset was generated from labels based on sensors placed at the chest, left ankle, and right wrist to detect upper body and limb movements. Data from each of the sensors were used to train the AI models in the FL process. Based on its superior overall classification performance, the heterogeneous global CNN model was trained with each of the three sensors. Evaluation of label classification performance shown in Fig.~\ref{fig:CNN_Performance} demonstrates that right wrist sensor data classified label activities with balanced accuracy close to 1.0. Exceptions to this were activities involving complete leg movements such as walking, climbing stairs, and cycling. Not surprisingly, the optimal placement of sensors to track human activities were the limbs, as the majority of physical activities involve hands and legs.

The state-of-art works baseline models in classifying physical activities were compared with proposed AI models. Except for the study by Anguita et al.~\cite{anguita2013public}, the other baseline models classified physical activities using deep learning CNN models. The state-of-works classified limited physical activities like walking, lying, sitting, walking upstairs, walking downstairs, standing, and jogging. The proposed design in this study was successful in classifying partial body motions like the frontal elevation of arms, waist bend forwards, and knees bending (crouching). Jiang et al.~\cite{jiang2015human} proposed a novel approach of assembling accelerometer and gyroscope signals as 2-D data as input to a deep CNN model for activity classification to reduce the computational cost. The secondary aim of this study was also met in that the optimum placement of sensors was determined. We have shown that a single sensor on the right wrist can perform better than~\cite{jiang2015human} achieving average balanced accuracy of 0.99 in classifying all 12 physical activities. 

\section{Conclusion}\label{conclusion}
Personalized monitoring is key to healthcare monitoring applications. This study focused on classifying individual client sensor data physical activities with various model architectures and ensembling the local models into a global robust model. The proposed decentralized FedStack architecture was able to outperform the state-of-art works and achieve better performance metrics. The study was able to identify and analyze the data collected from three sensors placed on a human body to classify full-body motion, partial-body motion, and still activities. The proposed design was evaluated by limiting one sensor input at a time to determine the optimum placement of sensors on the human body for activity recognition. This study overcomes the limitation in traditional federated architecture where clients might have differences in local model architectures and avoid model compilation problems in the global model. The limitations of the study are the global model is trained with all clients’ model predictions involved in the personalized monitoring. Also, the study assumes the same number of classification labels for all ten clients. Therefore, there might be differences in the number of labels across different clients. Other limitations would be the explainability of the AI models, which still is an issue to make informed decisions in domains like healthcare. From a technical perspective, the future direction of this study would be to explore and verify privacy-related issues in this proposed FedStack architecture. Scientifically, the study can be extended to have vital signs included for each client and classify them along with physical activities to enlarge the scope for an enhanced remote patient monitoring system.

\section*{Acknowledgment}

This research was partially funded by Queensland Health, Australia, and Grant DP220101360 from the Australian Research Council. 

\bibliographystyle{elsarticle-num}
\bibliography{KBS-refs}





\end{document}